\documentclass[acmtog,screen]{acmart}

\usepackage[utf8]{inputenc}
\usepackage[T1]{fontenc}
\usepackage{booktabs}
\usepackage[capitalise,noabbrev]{cleveref}
\crefformat{equation}{Equation~#2#1#3}
\graphicspath{{Sections/pics/}{./}}

\setcopyright{acmcopyright}
\acmJournal{TOG}
\acmYear{2019}
\acmVolume{38}
\acmNumber{6}
\acmArticle{178}
\acmMonth{11}
\acmDOI{10.1145/3355089.3356500}

\setcitestyle{square}

\newcommand{\norm}[1]{\left\lVert#1\right\rVert}
\newcommand{\domain}[1]{\mathcal{#1}}
\newcommand{\network}[1]{\mathrm{#1}}
\newcommand{\generator}[2]{\network{G}_{\domain{#1}\to\domain{#2}}}
\newcommand{\discrim}[1]{\network{D}_{\domain{#1}}}

\usepackage{tikz}
\usetikzlibrary{backgrounds,calc,arrows,arrows.meta,positioning}
\tikzset{annotate/.style={
		execute at begin scope={
			\pgfkeys{src/.store in=\src,src/.value required}
			\pgfkeys{pxwidth/.store in=\pxwidth,pxwidth/.value required}
			\pgfkeys{#1}
			\pgftransformshift{\pgfpointanchor{\src}{north west}}
			\pgfsetyvec{\pgfpointscale{1/\pxwidth}{\pgfpointdiff{\pgfpointanchor{\src}{north west}}{\pgfpointanchor{\src}{south west}}}}
			\pgfsetxvec{\pgfpointscale{1/\pxwidth}{\pgfpointdiff{\pgfpointanchor{\src}{north west}}{\pgfpointanchor{\src}{north east}}}}
}}}
\makeatletter\tikzset{add font/.code={\expandafter\def\expandafter\tikz@textfont\expandafter{\tikz@textfont#1}}}\makeatother
\tikzset{every node/.style={add font=\sffamily}}
\tikzset{every picture/.style={tight background,>=stealth}}

\begin{document}

\title{Neural Style-Preserving Visual Dubbing}

\author{Hyeongwoo Kim}
\affiliation{%
  \institution{Max Planck Institute for Informatics}
}
\email{hyeongwoo.kim@mpi-inf.mpg.de}

\author{Mohamed Elgharib}
\affiliation{%
  \institution{Max Planck Institute for Informatics}
}
\email{elgharib@mpi-inf.mpg.de}
	
\author{Michael Zollhöfer}
\affiliation{%
  \institution{Stanford University}
}
\email{zollhoefer@cs.stanford.edu}

\author{Hans-Peter Seidel}
\affiliation{%
  \institution{Max Planck Institute for Informatics}
}
\email{hpseidel@mpi-sb.mpg.de}

\author{Thabo Beeler}
\affiliation{%
  \institution{DisneyResearch|Studios}
}
\email{thabo.beeler@disneyresearch.com}

\author{Christian Richardt}
\orcid{0000-0001-6716-9845}
\affiliation{%
  \institution{University of Bath}
}
\email{christian@richardt.name}
	
\author{Christian Theobalt}
\affiliation{%
  \institution{Max Planck Institute for Informatics}
}
\email{theobalt@mpi-inf.mpg.de}

\renewcommand{\shortauthors}{Kim, Elgharib, Zollhöfer, Seidel, Beeler, Richardt and Theobalt}
	
\begin{abstract}
Dubbing is a technique for translating video content from one language to another.
However, state-of-the-art visual dubbing techniques directly copy facial expressions from source to target actors without considering identity-specific idiosyncrasies such as a unique type of smile.
We present a style-preserving visual dubbing approach from single video inputs, which maintains the signature style of target actors when modifying facial expressions, including mouth motions, to match foreign languages.
At the heart of our approach is the concept of motion style, in particular for facial expressions, i.e., the person-specific expression change that is yet another essential factor beyond visual accuracy in face editing applications.
Our method is based on a recurrent generative adversarial network that captures the spatiotemporal co-activation of facial expressions,
and enables generating and modifying the facial expressions of the target actor while preserving their style.
We train our model with unsynchronized source and target videos in an unsupervised manner using cycle-consistency and mouth expression losses, and synthesize photorealistic video frames using a layered neural face renderer. %
Our approach generates temporally coherent results, and handles dynamic backgrounds.
Our results show that our dubbing approach maintains the idiosyncratic style of the target actor better than previous approaches, even for widely differing source and target actors.
\end{abstract}



\begin{teaserfigure}
	\centering
	\includegraphics[width=\linewidth]{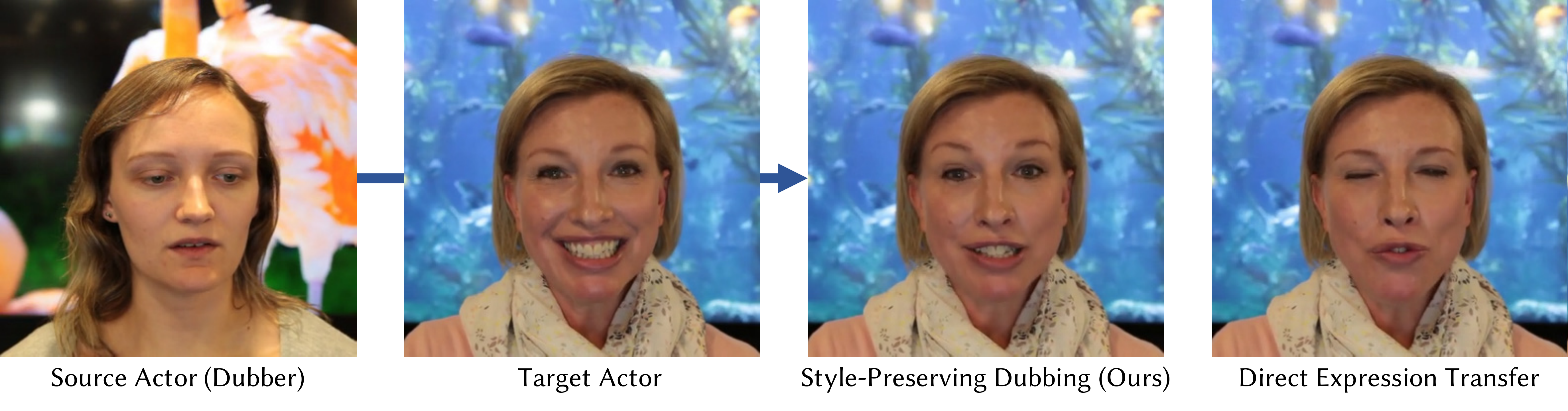}
	\caption{\label{fig:teaser}%
		Our visual dubbing method enables style-preserving lip synchronization by translating the source actor's facial expressions to a target actor's idiosyncratic style.
		Current dubbing techniques perform direct expression transfer from source to target actors.
		This reproduces the facial expressions of the source actor and leads to the loss of the style and idiosyncrasies of the target actor.
	}
\end{teaserfigure}

\maketitle


\section{Introduction}

Localization of media content, such as feature films or television series, has nowadays become a necessity since the target audience is oftentimes not familiar with the original spoken language.
The process to replace the original dialog with a different language, typically spoken by a different voice actor, is known as \emph{dubbing}.
The challenge that arises in traditional audio-only dubbing is that the audio and visual signals do not match anymore.
This is not only distracting but can significantly reduce understanding, since up to one third of speech information is captured from the visual signal in the presence of noise \cite{LeGoGCB1994}; and this is obviously aggravated for hearing-impaired viewers who rely on lip reading \cite{OwensB1985}.
Hence research (see \cref{sec:related}), and more recently industry (e.g., \url{https://synthesia.io/}), has started to address the problem of \emph{visual} dubbing, where the visual content is adjusted to match the new audio channel.

Every person speaks in a unique way, both in terms of expressions as well as their timing.
In particular, for actors, politicians and other prominent people, their idiosyncrasies and demeanor are part of their `brand' and it is of utmost importance to preserve their style when dubbing.
However, so far the field has entirely ignored style when dubbing.
This causes uncanny results, in particular for well-known actors – just imagine, for example, Robert DeNiro performing with the idiosyncrasies of Sylvester Stallone.

In this work, we propose the first method for visual dubbing that is able to preserve the style of the target actor, allowing to faithfully preserve a person's identity and performance when dubbing to new languages.
We achieve this by learning to automatically translate
a source performance to a target performance, requiring only unpaired videos of the two actors.
We learn how to perform the retargeting in parameter space using a cycle-consistency loss, and utilize a long short-term memory (LSTM) architecture to provide a solution that is temporally coherent.
We first convert the videos into our parametric representation using a multilinear face model and finally convert back to the video domain using a novel layer-based neural face renderer, which is capable of handling dynamic backgrounds.
More specifically, the novelties presented in this paper include:

\begin{itemize}
\item the first approach for visual dubbing that preserves the style of an original actor while transferring the lip movements according to a dubbing actor,
\item a novel target-style preserving facial expression translation network that we train in an unsupervised fashion using a temporal cycle-consistency loss and a mouth expression loss, and
\item a layer-based approach for neural face rendering that can handle dynamic video backgrounds.
\end{itemize}


\section{Related Work}
\label{sec:related}

Traditional dubbing pipelines seek to optimize the alignment between the dubbed audio and the mouth movements of the original target actor.
The source actor is recorded in a studio while reading out the dubbing script.
The script is translated in a way to maintain the overall semantics of the original script while trying to match the salient mouth movements of the target, such as the bilabial consonants /b/, /m/ and /p/.
The dubbing actor reads out the script and attempts to be in pace with the target's voice as much as possible.
Finally, the dubbed audio is manually edited to further improve its alignment with the target actor’s mouth region.
While commercial dubbing pipelines require professional dubbing actors and tedious manual editing, several techniques have been proposed to reduce the complexity of this process.
These techniques go back as far as the work of \citet{Brand1999} for voice puppetry.
The vast majority of dubbing-related techniques, however, can be divided into audio-based and visual-based approaches.

\subsection{Audio-Based Dubbing Techniques}

Audio-based dubbing techniques learn to associate the input audio stream of a driving source voice with the visual facial cues of a target actor.
This is challenging as there exists no one-to-one mapping between phonemes and visemes, i.e., the same sentence can be said with different expressions.
Motion-captured data are commonly used to represent facial visual cues, even in early work \cite{KshirM2003,DengN2006,MaCPWW2006,TayloMTM2012}.
Recently, the rise of deep learning has led to noticeable improvements in audio-based dubbing techniques.
\citet{KarraALHL2017} map the raw audio waveform to the 3D coordinates of a face mesh.
A trainable parameter is defined to capture emotions.
During inference, this parameter is modified to simulate different emotions.
\citet{TayloKYMKRHM2017} use a sliding-window deep neural network to learn a mapping from audio phoneme sequences to active appearance model (AAM) parameters.
The learned AAM parameters can be retargeted to different face rigs.
\citet{PhamCP2017} and \citet{ChaPWLRCQKSLISXFF2018} proposed a neural network approach to learn the mapping to facial expression blendshapes.
These approaches do not seek to generate photorealistic videos, and rather focus on controlling facial meshes or rigs for computer animation, or producing cartoon-looking characters.
Hence, they cannot be used for dubbing of more general visual content such as movies and TV shows.

The generation of photorealistic images or videos for audio dubbing has only seen little work to date.
\citet{ChungJZ2017} presented a technique that animates the mouth of a still image in a way that follows an audio speech.
Their approach builds on a joint embedding of the face and audio to synthesize the talking head.
\citet{VougiPP2018} introduce a similar method with a temporal generative model to produce more coherent speech-driven facial animation over time.
The end result is an animation of a still image and not yet a natural video.
\citet{SuwajSK2017} presented an audio-based dubbing technique with high-quality video animation as the end result.
A recurrent neural network is trained on 17 hours of President Obama's speeches to learn the mouth shape from the audio.
Their approach assumes the source and target have the same identity, and requires many hours of training data.
Hence, this approach cannot be applied directly to more general dubbing applications where source and target actors differ and data is relatively scarce.

\begin{figure*}
	\centering
	\includegraphics[width=\linewidth,trim=0 0 0 75,clip]{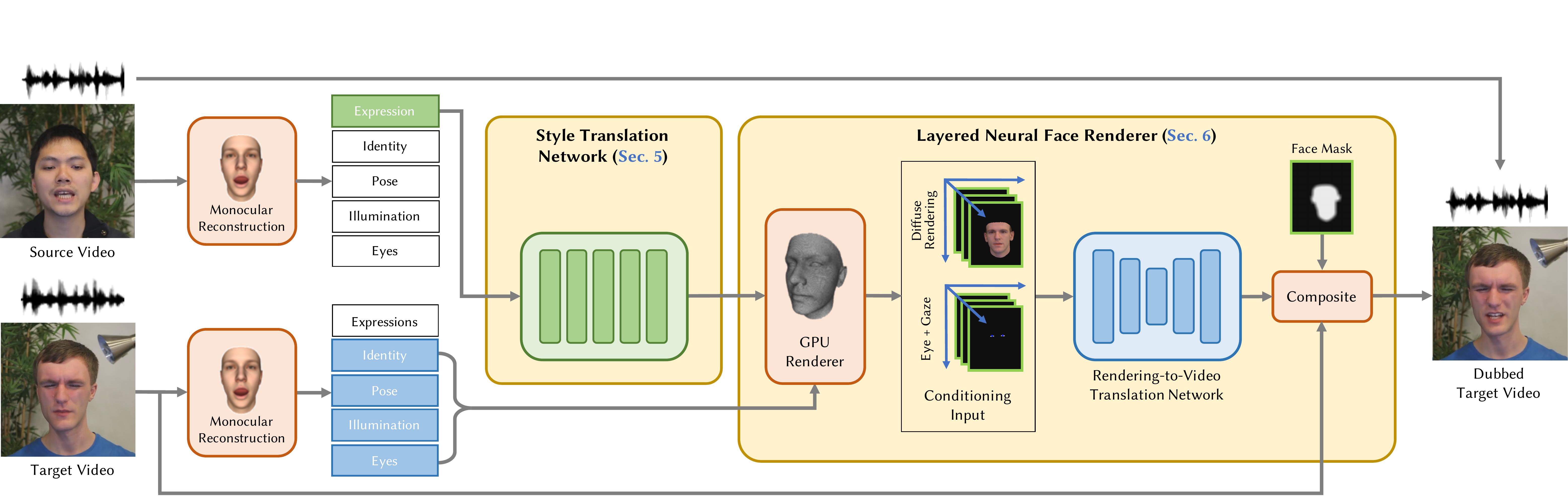}
	\caption{\label{fig:overview}%
		Overview of our style-preserving visual dubbing approach.
		From left to right:
		First, we reconstruct the parameters of a 3D face model from the source and target input videos (\cref{sec:model}).
		Next, we translate the source actor’s facial expressions using our novel target-style preserving recurrent generative adversarial network (\cref{sec:method}).
		Finally, we generate a photorealistic rendering of the dubbed target actor using a layer-based approach that composites a neural face rendering on top of dynamic video backgrounds (\cref{sec:renderer}).
	}
\end{figure*}

\subsection{Visual-Based Dubbing Techniques}

Visual dubbing techniques can be classified into image-based or model-based approaches.
Image-based techniques edit images directly in 2D image space.
\citet{GengSZWZ2018} presented a warp-guided technique capable of controlling a single target image through a source driving video.
The target image is warped according to the motion field of the driving video.
Two generative adversarial networks are used, one for adding photorealistic fine visual details, and the other for synthesizing occluded regions such as the mouth interior.
\citet{WilesKZ2018} presented X2Face, an approach for controlling a target video through a source video or audio.
They propose two networks: the first projects the target video into an embedded face representation, and the second network estimates a driving vector that encodes the desired facial expressions, head pose and so on.
While the approaches of \citeauthor{GengSZWZ2018} and \citeauthor{WilesKZ2018} produce compelling results, the outputs often suffer from unnatural movements.
In addition, neither approach is designed to maintain the target actor's style.

Model-based approaches rely on a parametric face model, namely a 3D deformable model.
The \citeauthor{BlanzV1999} face model \citeyearpar{BlanzV1999,BlanzBPV2003} is commonly used to represent the identity geometry and albedo of the face.
Facial expressions, including mouth movements, are usually modeled through blendshape parameters \cite{PighiHLSS1998}.
For dubbing, the facial expressions of the source video are directly copied to the target video.
\citet{GarriVSSVPT2015} transfer the blendshape weights of the mouth region from the source to the target video by overlaying a rendered target face model.
Finally, salient mouth movements, such as lip closure and opening, are imposed through the help of the dubbing audio track.
For this, the bilabial consonants (/b/, /m/ and /p/) are detected from the audio track.
\citeauthor{ThiesZSTN2016a}'s Face2Face \citeyearpar{ThiesZSTN2016a} allows dubbing in real time from a monocular source video by overlaying a modified rendered face model on the target.
Static skin texture is used and a data-driven approach synthesizes the mouth interior.
\citet{MaD2019} present an unpaired learning framework with cycle consistency for facial expression transfer.
Unlike our approach, it transfers the same mouth expression from the source actor without considering the target's style.
To this end, it introduces an additional lip correction term that simply measures the 3D distance of lip vertices.

\citet{KimGTXTNPRZT2018} presented Deep Video Portraits, a technique capable of producing high-quality photorealistic dubbing results.
At first, a synthetic rendering of the target actor is produced, which captures the facial expressions of the source actor while maintaining the target actor's identity and pose.
A conditional generative adversarial network translates the synthetic rendering into a photorealistic video frame.
This approach is trained per target video.
\citet{NaganSXWLSAFL2018} proposed a similar approach, which however does not require identity-specific training.
As a result, their approach can drive any still image by a given source video, but it only synthesizes the face region and not the hair.
The still image is assumed to have a frontal perspective and a neutral pose.
Even though model-based techniques provide full control over the target video, 
many suffer from audio-visual misalignments and are not designed to create high-quality videos with general backgrounds.
In addition, model-based techniques often exhibit noticeable artifacts in synthesized mouth interiors \cite{GarriVSSVPT2015,ThiesZSTN2016a,NaganSXWLSAFL2018} or dynamic skin textures \cite{ThiesZSTN2016a}.
Moreover, none of them maintain the style of the target actor.

\subsection{Image-to-Image Translation}

Learning-based image-to-image translation techniques have shown impressive results for a number of applications \cite[e.g.][]{IsolaZZE2017}.
The core component is a conditional generative adversarial network that learns a mapping from the source to the target domain.
This, however, requires paired training data, an assumption that is not easily satisfied.
The introduction of cycle-consistency losses enabled learning from unpaired training data \cite{ZhuPIE2017,KimCKLK2017,YiZTG2017} without explicit training pairs of source and target images.
The cycle-consistency loss is defined such that a mapping from the source to the target followed by the inverse of this mapping should lead to the original source.
This constraint is also applied in the opposite direction separately, i.e., from the target to the source.
In our work, we utilize a cycle-consistency loss in model parameter space to train our visual dubbing framework on unsynchronized training data.


\section{Overview}

Different actors speak in different ways, using their own facial expressions.
These person-specific idiosyncrasies need to be preserved during the visual dubbing process, hence one cannot simply copy the facial expressions from a source actor to a target actor as done in previous work.
In order to achieve this, we propose to learn a \emph{style-preserving} mapping between facial expressions in an unsupervised manner.

Our approach consists of three stages (\cref{fig:overview}): monocular face reconstruction, style-preserving expression translation, and layered neural face rendering.
The first stage of our approach registers a 3D face model to the source and target input videos (\cref{sec:model}).
This step reconstructs the facial expression parameters for every video frame.
The second stage of our approach is a novel style-preserving translation network (\cref{sec:method}).
We introduce a recurrent generative adversarial network that learns to transfer the source actor's expressions while maintaining the idiosyncrasies of a specific target actor.
We train this network in an unsupervised manner on unpaired videos using cycle-consistency and mouth expression losses.
The third stage of our approach is a new layered neural face renderer (\cref{sec:renderer}) that generates photorealistic video frames from the style-translated expression parameters.
We adopt the recent neural face rendering approach of \citet{KimGTXTNPRZT2018}, and extend it for dynamic video backgrounds.
Specifically, we introduce a soft face mask to blend the rendered photorealistic faces with the existing target video background layer.


\section{3D Face Modeling}
\label{sec:model}

We map the expression transfer problem from screen space to parameter space by registering a parametric 3D face model to every video frame.
This later enables us to robustly learn a style-preserving expression mapping from the source to the target actor domains in an unsupervised manner.

We employ a parametric face model that encodes the head pose, face identity (geometry and appearance), and facial expression based on a low-dimensional vector.
In more detail, we recover, for each frame $f$, the pose of the head $\mathbf{T} \!\in\! \mathrm{SE}(3)$, face geometry $\boldsymbol\alpha \!\in\! \mathbb{R}^{80}$, face reflectance $\boldsymbol\beta \!\in\! \mathbb{R}^{80}$, face expression $\boldsymbol\delta \!\in\! \mathbb{R}^{64}$, and spherical-harmonics illumination $\boldsymbol\gamma \!\in\! \mathbb{R}^{27}$.
For dimensionality reduction, the geometry and appearance bases have been computed based on 200 high-quality scans \cite{BlanzV1999} using principal component analysis (PCA).
The low-dimensional expression subspace has been computed via a PCA of the facial blendshapes of \citet{CaoWZTZ2014} and \citet{AlexaRLCMWD2010}.
We recover all parameters from monocular video based on an optimization-based 3D face reconstruction and tracking approach inspired by \citet{GarriZCVVPT2016} and \citet{ThiesZSTN2016a}.
The energy is composed of a dense color alignment term 
between the input image and the rendered model, a sparse alignment term 
based on automatically detected facial landmarks \cite{SaragLC2011}, and a statistical regularizer.
The facial landmark tracker also recovers the 2D image position of the pupils of the left eye, $\mathbf{e}_\text{l} \!\in\! \mathbb{R}^2$, and right eye, $\mathbf{e}_\text{r} \!\in\! \mathbb{R}^2$.
This procedure lets us fully automatically annotate each video frame $f$ with a low-dimensional parameter vector $\mathbf{p}_f \!\in\! \mathbb{R}^{261}$.
Of specific importance for us in the later processing steps are the recovered expression parameters $\boldsymbol\delta$.
In the following, we show how to robustly learn a style-preserving mapping between the expression parameters of two different actors without requiring paired training data.


\section{Style Translation Network}
\label{sec:method}

We propose a style translation network that learns a mapping from the distribution of the source actor expressions to the distribution of the target actor expressions, and vice versa, using cycle consistency.
Our approach is inspired by recent techniques for unpaired image-to-image translation \cite[e.g.][]{ZhuPIE2017,KimCKLK2017,YiZTG2017}, and shares a similar high-level design:
We employ a generative adversarial network with two generator networks and two discriminator networks, which are trained in an unsupervised fashion from unpaired training data using cycle-consistency, adversarial and mouth expression losses.
The generator networks translate from the source to the target domain, and vice versa, while there is a discriminator network for each of the two domains.
However, everything else is different as we learn the translation of temporal facial expression parameters, i.e., multiple vectors $\boldsymbol\delta$ corresponding to multiple video frames.
Specifically, we propose a \emph{recurrent} generative adversarial network to encode the temporal dynamics of the learned distribution of facial expressions using long short-term memory (LSTM) units \citep{HochrS1997}.

\subsection{Network Architecture}
\label{sec:architecture}

\begin{figure}
	\centering
	\includegraphics[width=\linewidth]{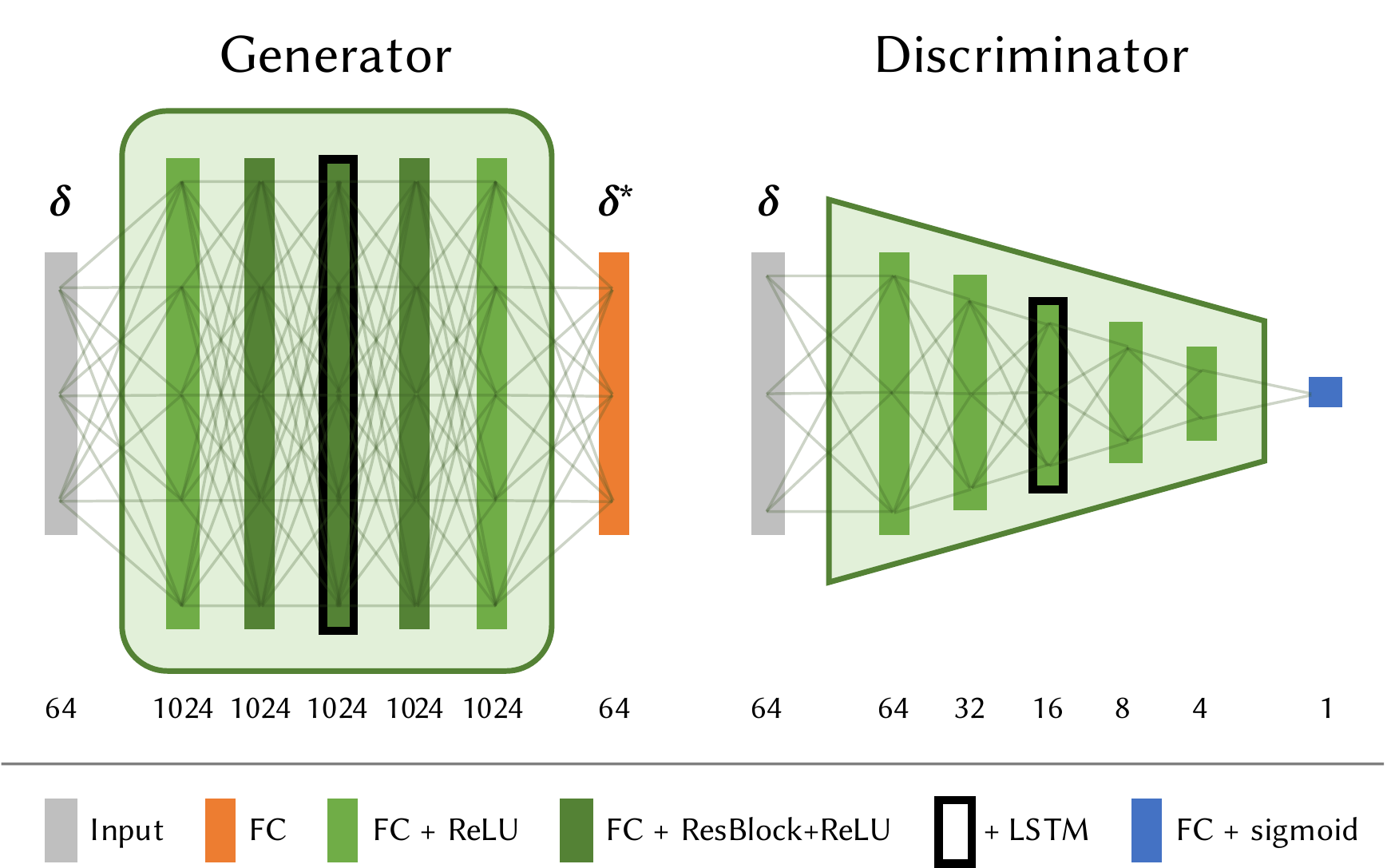}
	\caption{\label{fig:network2-TSP}%
		Illustration of the architectures of the generator and discriminator networks as part of our style translation network (\cref{sec:method}).
	}
\end{figure}

The input to both the generator and the discriminator networks is a tuple $(\boldsymbol\delta_{t - N + 1}, \ldots, \boldsymbol\delta_{t})$ of $N \!=\! 7$ facial expression parameters $\boldsymbol\delta_f \!\in\! \mathbb{R}^{64}$.
The tuple comprises the six frames before frame $t$ and the current frame $t$.
Each parameter vector $\boldsymbol\delta$ is calculated using the approach in \cref{sec:model}, and we normalize each expression component to zero mean and unit variance (per video).
We illustrate the architecture of our networks in \cref{fig:network2-TSP}.

\paragraph{Generators}

The generator consists of five fully-connected layers, each containing 1024 nodes with ReLU activations.
We use residual blocks \citep{HeZRS2016} in the middle three layers to facilitate the learning of deviations from the identity expression translation function.
This effectively captures the difference between the target and source distributions.
We further implement LSTM units in the middle layer of the generator to encode the temporal dynamics of facial expressions.
This is defined over $N \!=\! 7$ consecutive frames, specifically the current frame and the six preceding frames.
A final fully-connected layer of 64 nodes outputs the parameters of the translated expression coefficients, without any activation function.

\paragraph{Discriminators}

The discriminator network comprises five fully-connected layers with ReLU activations.
The first layer has 64 nodes, matching the size of the input, and each subsequent layer has half as many nodes.
The middle layer also implements LSTM units to consider the temporal dynamics of facial expressions.
The output node is fully connected to the previous layer and uses a sigmoid activation to produce a value in the unit range.
Given a tuple of $N$ facial expression vectors as input, the discriminator produces an output with $N$ real values in the unit range.

\subsection{Training Loss}

To train our style translation network in an unsupervised manner, we combine three losses into our objective function:
\begin{align}\label{eq:loss}
L = \lambda_\text{cc} L_\text{cc} + \lambda_\text{adv} L_\text{adv} + \lambda_\text{me} L_\text{me} \text{.}
\end{align}
Here, $L_\text{cc}$ is the cycle-consistency loss that enables training with unpaired training data, $L_\text{adv}$ is the adversarial loss that encourages the output of the generator to better match the target domain, and $L_\text{me}$ is a novel cosine mouth expression loss that promotes corresponding mouth expressions, such as mouth closure, between source and target.
Each loss is weighted by a corresponding coefficient $\lambda_\bullet$.

\subsubsection{Cycle-Consistency Loss}

There are two generators in our network that translate facial expression parameters from the distribution of source actor expressions, $\domain{S}$, to the distribution of target actor expressions, $\domain{T}$, and vice versa.
$\generator{S}{T}$ denotes the translation of the source actor expression $\mathbf{s} \!=\! (\boldsymbol\delta_{t - N + 1}, \ldots, \boldsymbol\delta_{t}) \!\in\! \domain{S}$ to a target actor expression $\generator{S}{T}(\mathbf{s}) \!\in\! \domain{T}$.
This captures the spatial and temporal co-activations of the facial expressions while preserving the style of the target.
On the other hand, $\generator{T}{S}$ is the mapping in the opposite direction, from the target distribution $\mathbf{t} \!\in\! \domain{T}$ into the source distribution $\generator{T}{S}(\mathbf{t}) \!\in\! \domain{S}$.
Composing both generators, and measuring the distance to the starting point, results in the cycle-consistency loss
\begin{align}\label{eq:cc-loss}%
L_\text{cc} = \norm{ \generator{T}{S}(\generator{S}{T}(\mathbf{s})) - \mathbf{s}}_{1}
            + \norm{ \generator{S}{T}(\generator{T}{S}(\mathbf{t})) - \mathbf{t}}_{1} \text{,}
\end{align}
where $\mathbf{s} \!\in\! \domain{S}$ and $\mathbf{t} \!\in\! \domain{T}$ are unpaired training samples, and we use an $\ell_1$-loss to measure similarity.
Using both generators with cycle consistency allows us to train our approach in an unsupervised manner, with unpaired data.
This is important since it is challenging to obtain paired data (time-synchronized face video across different languages) for our problem of visual dubbing from one language to another.

\subsubsection{Adversarial Loss}

Both generators are accompanied by discriminators, $\discrim{S}$ and $\discrim{T}$, which correspond to the source and target domains, respectively.
The discriminators work towards getting better in classifying the generated result as either real or synthetic, while the generators aim to fool the discriminators by improving the quality of their output.
The input to each discriminator is a temporal vector of $N$ facial expression vectors, and its output is a vector of $N$ real numbers, corresponding to the individual input vectors.
This, in combination with the LSTM units, allows us to better capture the temporal correlations between the examined distributions.
We define the adversarial loss in a bidirectional manner as follows:
\begin{align}\label{eq:adv-loss}%
\begin{split}
L_\text{adv} &      = \log \frac{\norm{ \discrim{T}(\mathbf{t}) }_1}{N} + \log\!\left( 1 - \frac{\norm{\discrim{T}(\generator{S}{T}(\mathbf{s})}_1}{N} \right) \\
             & \;\! + \log \frac{\norm{ \discrim{S}(\mathbf{s}) }_1}{N} + \log\!\left( 1 - \frac{\norm{\discrim{S}(\generator{T}{S}(\mathbf{t})}_1}{N} \right) \text{.}
\end{split}
\end{align}
As before, $\mathbf{s} \!\in\! \domain{S}$ and $\mathbf{t} \!\in\! \domain{T}$ are unpaired training samples.

\subsubsection{Cosine Mouth Expression Loss}

The unpaired training using cycle-consistency and adversarial losses does not always preserve important mouth expressions, such as opening or closing.
We therefore introduce an additional loss to encourage the correct translation of these important mouth expressions.
Specifically, we use a cosine loss on the ten mouth-specific facial expressions between the source and target domains.
The cosine loss is more effective in aligning the mouth-related source and target expressions with different magnitudes, corresponding to different styles, than the Euclidean loss.
This encourages our network to maintain correspondence in mouth expressions without unduly constraining the strength of these expressions, as this would for example alter the intensity of smiles between actors.
We again use a symmetric loss,
\begin{align}\label{eq:me-loss}
L_\text{me} = L_\text{cos}(\mathbf{s}, \generator{S}{T}(\mathbf{s}))
            + L_\text{cos}(\mathbf{t}, \generator{T}{S}(\mathbf{t})) \text{,}
\end{align}
where $L_\text{cos}$ computes the mean cosine distance between the mouth-specific expressions over all time steps:
\begin{align}\label{eq:cos-loss}
L_\text{cos}(\mathbf{s}, \mathbf{t}) =
  \frac{1}{N} \sum_{n=1}^{N}
  \frac{
    \boldsymbol\mu(\mathbf{s}_n) \cdot
    \boldsymbol\mu(\mathbf{t}_n)
  }{
    \norm{\boldsymbol\mu(\mathbf{s}_n)}_2 \cdot
    \norm{\boldsymbol\mu(\mathbf{t}_n)}_2
  } \text{.}
\end{align}
Here, we use the notation $\mathbf{s}_n$ to select the $n^\text{th}$ element of the tuple $\mathbf{s}$, which is the facial expression vector $\boldsymbol{\delta}$ of a source actor, and from which the function $\boldsymbol\mu(\cdot)$ selects the ten mouth-specific expression coefficients.
We select the ten coefficients with the largest mouth expression variation by visual inspection of the rendered PCA basis.

\subsection{Network Training}

Our training dataset is a collection of sequential expression parameters from individual videos, recovered using the monocular 3D face reconstruction approach in \cref{sec:model}.
We found that facial expression styles consistently captured by approximately five-minute-long videos are typically sufficient to train our style translation network. 
As preprocessing, we normalize each expression coefficient to zero mean and unit variance, and then extract sliding windows of size $N \!=\! 7$ frames.
We balance the influence of loss functions in \cref{eq:loss} using $\lambda_\text{cc} \!=\! 10$, $\lambda_\text{adv} \!=\! 1$ and $\lambda_\text{me} \!=\! 5$.
The loss is minimized using the Adam solver \citep{KingmB2015} with an initial learning rate of 0.0001 and an exponential decay rate of 0.5. 
During backpropagation for the LSTM units, we apply a gradient norm clipping operation to avoid exploding gradients \citep{PascaMB2013}.
We implement our network using the TensorFlow deep learning library \citep{AbadiABBCCCDDDGGHIIJJKKLMMMMOSSSSTTVVVVWWWYZ2015}; training typically converges within 25 epochs.


\section{Neural Face Renderer}
\label{sec:renderer}

The final step of our approach is to synthesize a photorealistic portrait video from a sequence of face model parameters (see \cref{sec:model}) that correspond to the dubbed target actor.
Our approach builds on recent advances in neural rendering for creating high-fidelity visual dubbing results.
Specifically, we extend deep video portraits \cite{KimGTXTNPRZT2018}, which assumes a static video background, to support the dynamic video backgrounds found in feature films, television series, and many real-world videos.
Considering our target application of dubbing, the only area that needs to be modified is the face interior of the target actor, as we would like to preserve the remainder of the target actor's performance.
Unlike \citeauthor{KimGTXTNPRZT2018}, we therefore restrict the neural renderer to the face region and composite the predicted face rendering over the target video.
This leaves any potentially dynamic background intact in the final face rendering.

\begin{figure}
	\centering
	\includegraphics[width=\columnwidth,trim=0 30 0 30,clip]{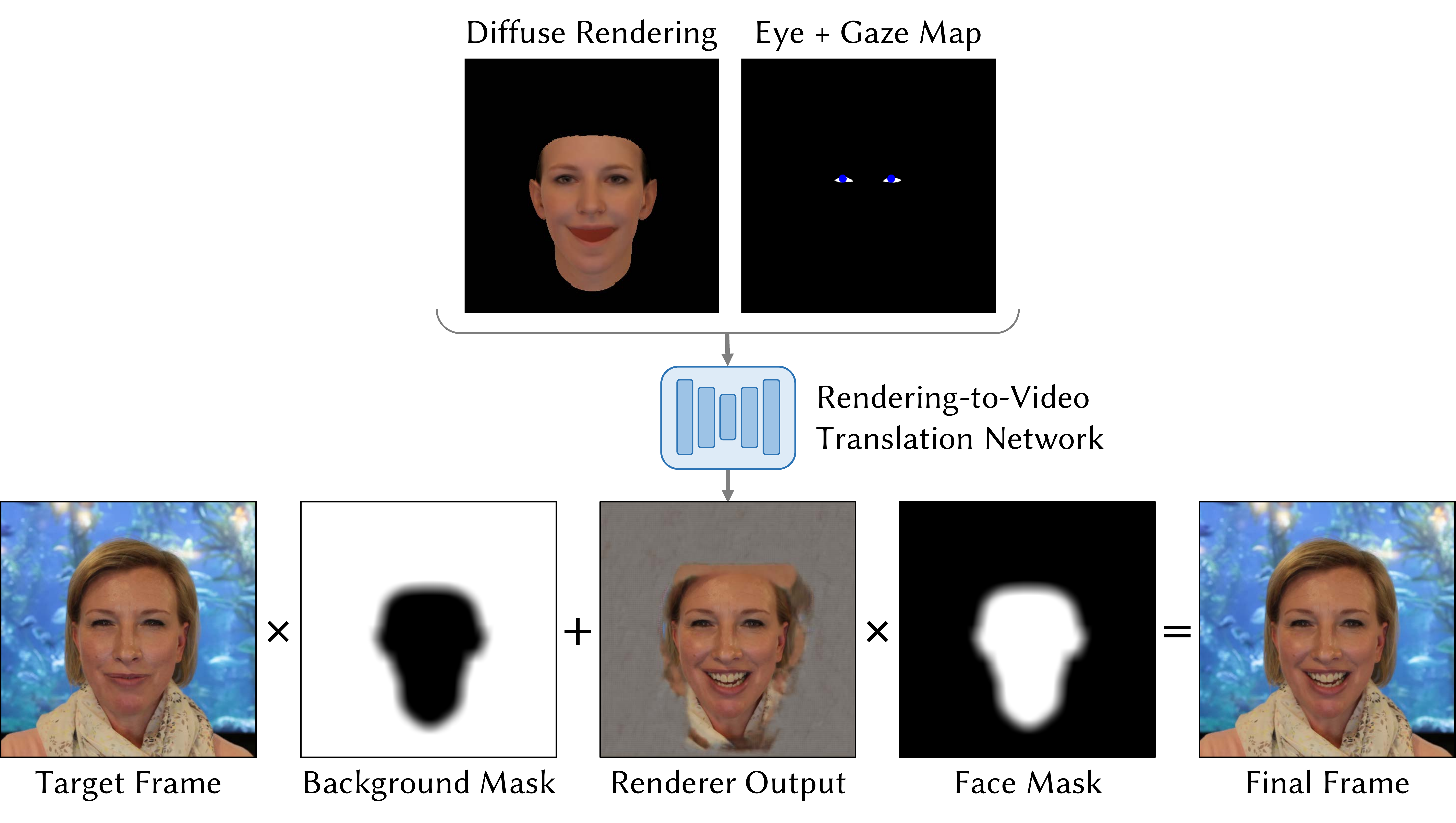}
	\caption{%
		We use a layer-based neural face renderer that composes the neural face rendering onto the dynamic video background using a soft face mask.
	}
	\label{fig:renderer}
\end{figure}

\begin{figure*}[t]
	\centering
	\input{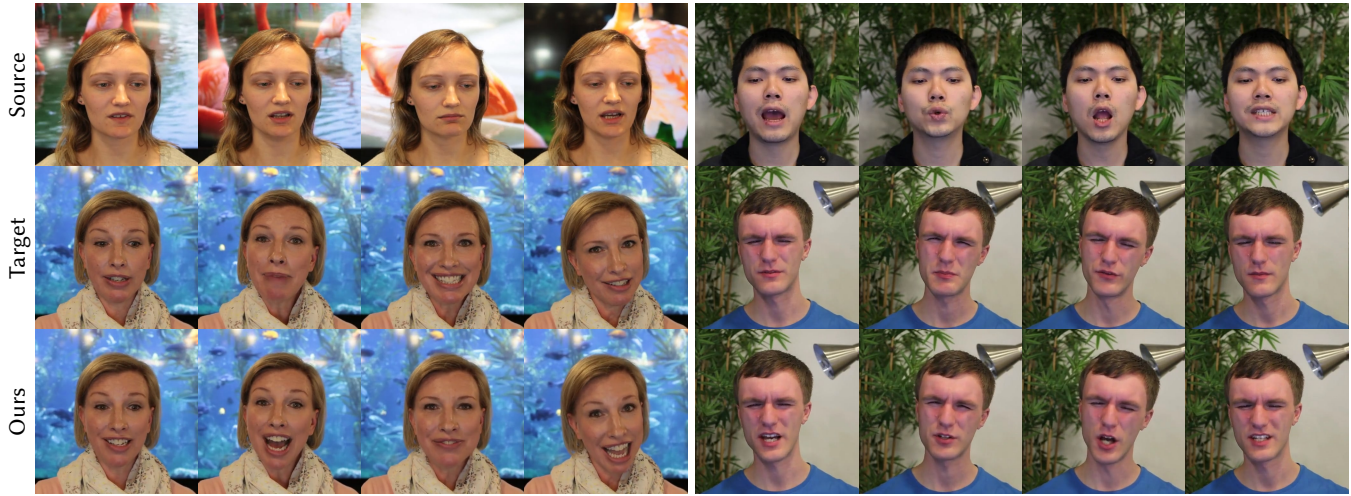}
	\caption{\label{fig:result-dubbing}%
		We demonstrate our style-preserving visual dubbing approach for two German-speaking `Target' actors with strong expressive styles (in the middle row), who we are dubbing into English using neutral `Source' actors (in the top row).
		As can be seen, our results (in the bottom row) do preserve the idiosyncrasies and the style of the target actors well.
		Note that all videos feature dynamic backgrounds.
	}
\end{figure*}

\Cref{fig:renderer} shows the work flow of our layer-based renderer.
We first rasterize diffusely shaded renderings of the face model, as well as eye maps with proxy pupils using the standard graphics pipeline.
Unlike \citet{KimGTXTNPRZT2018}, we do not render texture coordinates as we found them to be not necessary.
We feed these images into \citeauthor{KimGTXTNPRZT2018}’s rendering-to-video translation network \citeyearpar{KimGTXTNPRZT2018}, which we use at 512$\times$512\,pixel resolution by default.
To focus the network on the face region, we apply the face mask to the predicted image and the ground truth before passing them to the discriminator and computing the adversarial loss.
The employed additional per-pixel $\ell_1$-loss with respect to ground truth is only computed within the face mask region, and the relative weight between the $\ell_1$ and adversarial losses is set to a ratio of 100:1, as in \citet{KimGTXTNPRZT2018}.
The face mask covers the face interior between the ears, from the forehead down to the laryngeal prominence (Adam's apple).
Finally, we composite the predicted face over the current target video frame.
To achieve seamless blending, we erode the binary face mask to reduce its size slightly, and then smooth its boundaries with a Gaussian filter.
The face rendering is then composited onto the original target video using the soft face mask.
Since we did not modify the head pose, the composition appears seamless in most cases.
For better temporal consistency in the generated results, we process a video using a moving window.
The input to our network is a space-time tensor defined over 7 frames (including the current frame as the last frame).
The tensor therefore has a dimension of $W \!\times\! H \!\times\! (7\!\cdot\!2\!\cdot\!3)$, i.e., stacking all 7~conditioning inputs (2~images with 3~color channels each).


\section{Results}

We demonstrate our style-preserving video dubbing approach, perform a qualitative and quantitative evaluation (user study), and thoroughly compare to the state of the art in audio-based and video-based dubbing.
Please see our supplemental video for audio-visual results and comparisons.
We start by giving an overview of the used sequences, and discuss the runtime requirements of our approach.

\paragraph{Datasets}

We tested our approach on a diverse set of 11 source and 12 target sequences, which are detailed in \cref{tab:datasets}.
The average length of both the source and target sequences is on average five minutes.
In total, we dubbed over 50 minutes of video footage with our approach.
The source and target sequences show different people, all with their own person-specific idiosyncrasies and style, in front of a large variety of backgrounds, both static as well as dynamic.
We also show dubbing results between different languages, such as German-to-English dubbing.
The resolution of all produced videos is 512$\times$512\,pixels.

\paragraph{Runtime Requirements}

Dense 3D face reconstruction and tracking takes 250\,ms per video frame.
Training our style-preserving expression mapping takes 8\,hours per sequence on an Nvidia Tesla V100.
At test time, applying the mapping takes 952\,ms per video frame and our neural renderer requires additional 224\,ms to produce the final output.

\paragraph{Training and Testing}

We learn the style-preserving mapping between the source and target sequences in an unsupervised manner.
An input video is split into training and test sets.
We usually use the first 7,500 frames for training.
At test time, we feed the the rest of the frames to our system as illustrated in \cref{fig:overview}.
Our style translation network is source-to-target specific.
Therefore, we retrain the network with different video pairs to handle different styles or identities.
Our neural renderer is also person-specific and trained with all the frames from a target video without any split.

\begin{figure*}[t]
	\centering
	\input{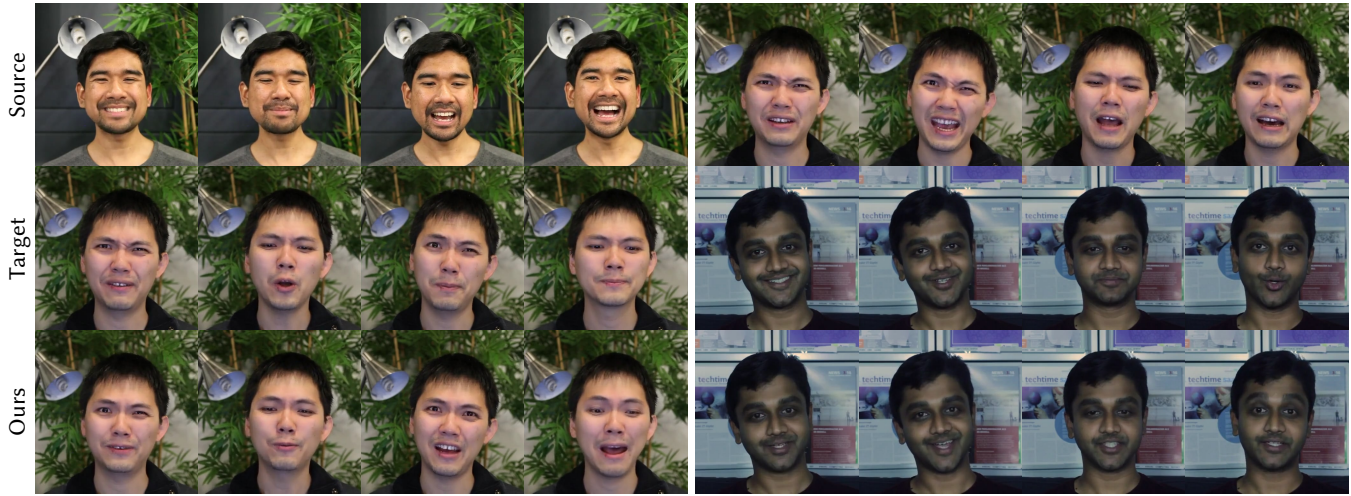}
	\caption{\label{fig:result-exp}%
		Our dubbing technique can handle expressive source styles.
		As can be seen, our approach is able to transfer the source facial expressions, while preserving the idiosyncrasies and style of the target actor.
		We demonstrate English-to-Indonesian dubbing (left) and Kannada-to-English dubbing (right).
	}
\end{figure*}

\subsection{Visual Dubbing}

\begin{figure*}[p]
	\centering
	\input{Sections/pics/result-similar-expressions-v3.tex}
	\caption{\label{fig:result-similar-expressions}%
		Our dubbing technique naturally supports source and target actors with similar expressions.
		In these examples, both actors are either smiling (left) or neutral (right).
		In each case, our dubbing result (bottom row) maintains the specific style of each target actor.
		On the left, the target has an expressive smile with a wide mouth opening, which is maintained in our result.
		On the right, the target actor speaks in a narrower mouth shape than the source.
		This target style is preserved in our result. 
		For these and additional video results, please refer to the supplemental video.
	}
\vspace{1.5em}
	\centering
	\input{Sections/pics/comparison-dvp-merged-v2.tex}
	\caption{\label{fig:comparsion_dvp}%
		Comparison to a variant of Deep Video Portraits \citep{KimGTXTNPRZT2018}, i.e. our approach without the style translation network.
		Our approach enables us to preserve the style and idiosyncrasies of each target actor.
		This is in contrast to other dubbing approaches, such as \citet{KimGTXTNPRZT2018}, that copy expressions directly and are hence not style-preserving.
		On the left, our results maintain the wide-open mouth of the target, while the variant of \citeauthor{KimGTXTNPRZT2018} narrows it to match the source style.
		On the right, our technique captures the target style more naturally, while the variant of \citeauthor{KimGTXTNPRZT2018} generates artifacts (see arrows).
	}
\end{figure*}

Visual dubbing is an approach to change the mouth motion of a target actor, such that it matches the voice of a dubbing actor that speaks in a foreign language.
One example of this is dubbing an English movie to German.
Existing visual dubbing approaches directly copy the mouth motion of the dubbing actor to the target.
While this leads to good audio-visual alignment, it also removes the person-specific idiosyncrasies and the style of the target actor, and makes the target actor's face move unnaturally like the source actor.
Our style-preserving video dubbing approach enables to achieve good audio-visual alignment, while also preserving the idiosyncrasies and the style of the target actor, see \cref{fig:result-dubbing}.
For example, we dub a very expressive actor speaking in German to English based on a neutrally speaking dubbing actor.
As can be seen, our approach is able to preserve the idiosyncrasies and the style of the target actor well.
For the female target actor (\cref{fig:result-dubbing}, left), our technique maintained her wide and happy mouth openings.
For the the male target actor (\cref{fig:result-dubbing}, right), we maintained his near-closed eyes while modifying his mouth movements to match the dubbing track.
For the full result videos, we refer to the supplemental video.

Our style-preserving dubbing approach can also be used to handle cases when the source actor has an expressive style.
This is a more challenging problem as the translation network needs to remove the strong source style and replace it with the different target style.
\Cref{fig:result-exp} shows the results from expressive styles.
On the left, the source actor is smiling while the target is angry.
On the right, the source actor is angry while the target is smiling.
In both examples, our approach captures the source mouth movements and maintain the target style.
These examples also demonstrate our visual dubbing method in other languages: English to Indonesian on the left and Kannada to English on the right.

Our dubbing technique naturally handles similar expressions between the source and target actors, as we show in \cref{fig:result-similar-expressions}.
We note that similar expressions still appear differently due to person-specific styles.
Our technique dubs the target actor while maintaining his/her specific style.
We include additional results in our supplemental video.

\subsection{Comparisons to the State of the Art}

We perform extensive comparisons to the current state of the art in visual dubbing, facial reenactment and audio-based reenactment.
More specifically, we compare to the VDub \citep{GarriVSSVPT2015}, Deep Video Portraits \citep{KimGTXTNPRZT2018}, and Audio2Obama \citep{SuwajSK2017} approaches.
We also perform a baseline comparison with the unpaired image-to-image translation approaches CycleGAN \citep{ZhuPIE2017} and UNIT \cite{LiuBK2017}.

\begin{figure}
	\centering
	\input{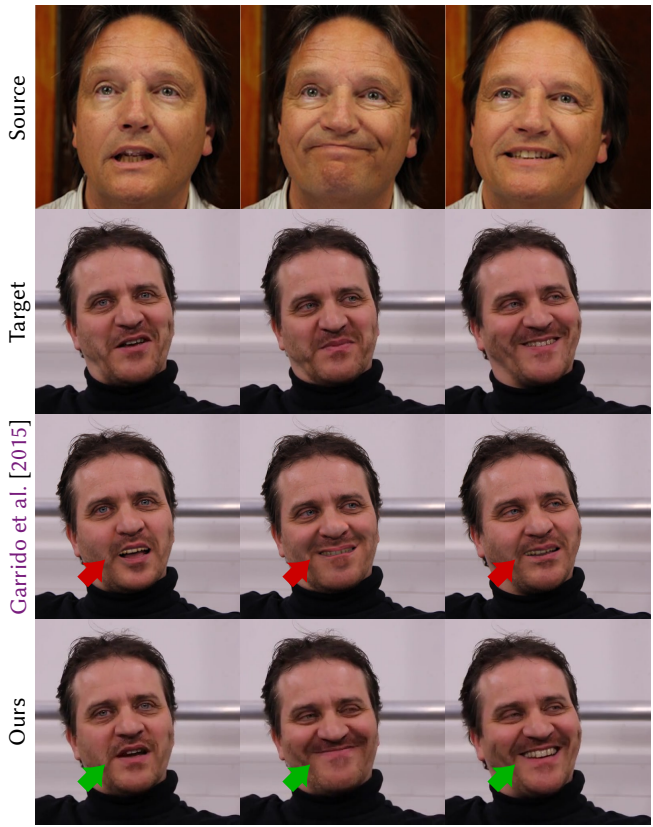}
	\caption{\label{fig:comparsion_vdub}%
		Comparison to \citet{GarriVSSVPT2015}.
		Our approach is able to preserve the style and idiosyncrasies of the target actor, while these get lost with the approach of \citet{GarriVSSVPT2015}.
		In addition, our approach synthesizes a higher quality and more realistic mouth interior.
	}
\end{figure}

First, we compare to the state-of-the-art audio-visual VDub approach \citep{GarriVSSVPT2015} in \cref{fig:comparsion_vdub}.
The VDub approach leverages the audio channel to better align the visual content with lip closure events, but is not able to preserve the style of the target actor.
In contrast, our style-preserving visual dubbing approach enables us to maintain the style of the target actor and achieve a good audio-visual alignment.
In addition, our learning-based approach synthesizes a higher quality mouth interior than the model-based VDub approach that only renders a coarse textured teeth proxy.

In \cref{fig:comparsion_dvp}, we compare our approach to a variant of Deep Video Portraits \citep{KimGTXTNPRZT2018}, a learning-based facial reenactment approach that also supports visual dubbing.
Specifically, we use our approach without the style translation network, i.e., we pass the source facial expressions directly to our layer-based renderer, which handles dynamic backgrounds unlike Deep Video Portraits.
Unlike other dubbing approaches, our style-preserving visual dubbing approach enables us to maintain the style and idiosyncrasies of the target actor.
For more detail, please refer to the supplemental video.

\begin{figure}
	\centering
	\input{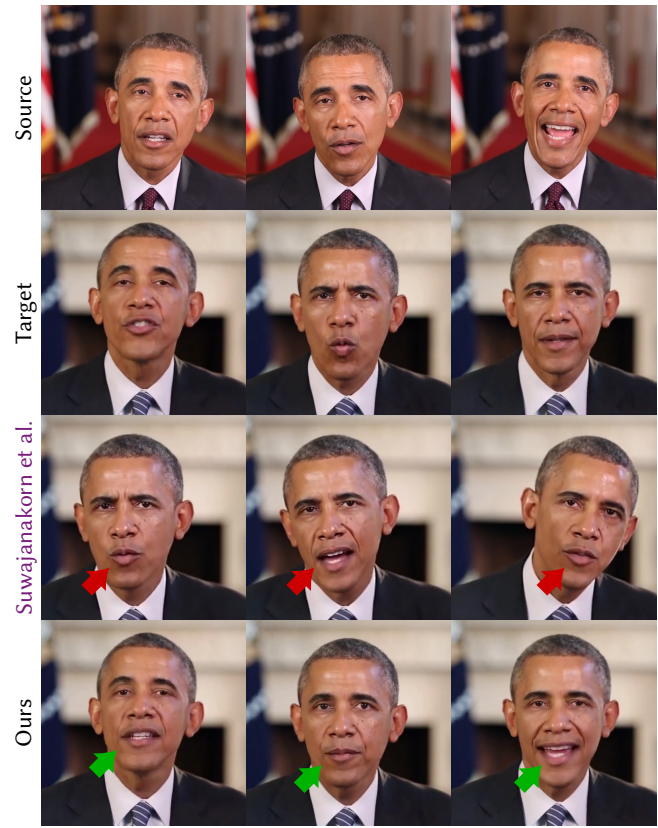}
	\caption{\label{fig:comparsion_synthobama}%
		Comparison to the audio-based dubbing approach of \citet{SuwajSK2017}.
		Their results are not always consistent with the source actor, with the mouth being open or closed when it should not be (see red arrows).
		In addition, this approach does not explicitly deal with style, as it can only perform self-reenactment.
		Videos: The White House (public domain).
	}
\end{figure}

We compare to the audio-based Audio2Obama facial reenactment approach \citep{SuwajSK2017} in \cref{fig:comparsion_synthobama}.
Their approach can control a virtual version of Barack Obama based on a new audio clip of himself.
Note that since this approach only works from-Obama-to-Obama, it does not explicitly deal with style, since there is no style to preserve in this setting.
Nevertheless, our visual dubbing approach leads to a more faithful reproduction of the actual visual content in the dubbing sequence, while the audio-based approach misses mouth motions that are uncorrelated with the audio track.
In addition, Audio2Obama is tailored for Barack Obama and trained on 17 hours of his speeches.
Our technique, however, is trained on only 2 minutes from just a single video.
For video results, please see our supplemental video.

\begin{figure}
	\centering
\begin{tikzpicture}[
image/.style={inner sep=0pt, outer sep=0pt},
collabel/.style={above=9pt, anchor=north, inner ysep=0pt, scale=0.9, align=center},
rowlabel/.style={left=17pt, rotate=90, anchor=north, inner ysep=0pt, scale=0.9, align=center},
subcaption/.style={inner xsep=0.75mm, inner ysep=0.75mm, below right},
arrow/.style={-{Latex[length=2.5mm,width=4mm]}, line width=2mm},
]

\def\padding{0pt}
\newcommand{\subfig}[2][0px 0px 0px 0px]{\includegraphics[width=0.232\linewidth, trim=#1, clip]{comparison-cyclegan-v2/#2.jpg}}

\node [image]                (static-source)   at (0,0)                  {\subfig{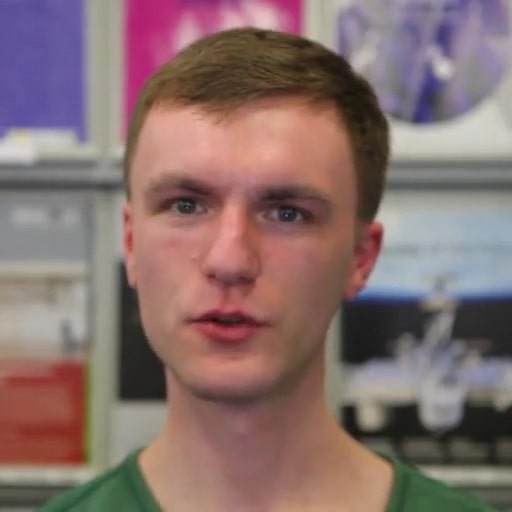}};
\node [image,right=\padding] (static-cyclegan) at (static-source.east)   {\subfig{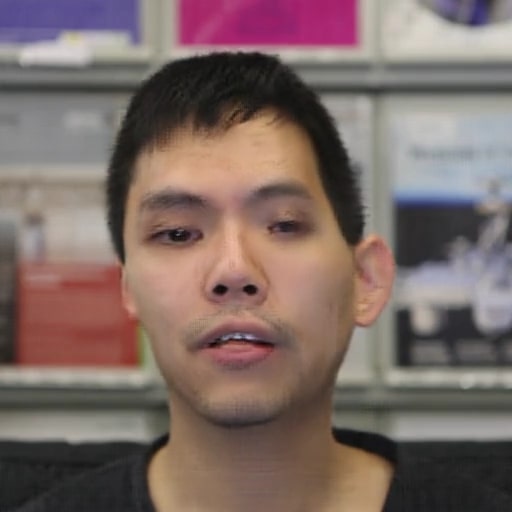}};
\node [image,right=\padding] (static-unit)     at (static-cyclegan.east) {\subfig{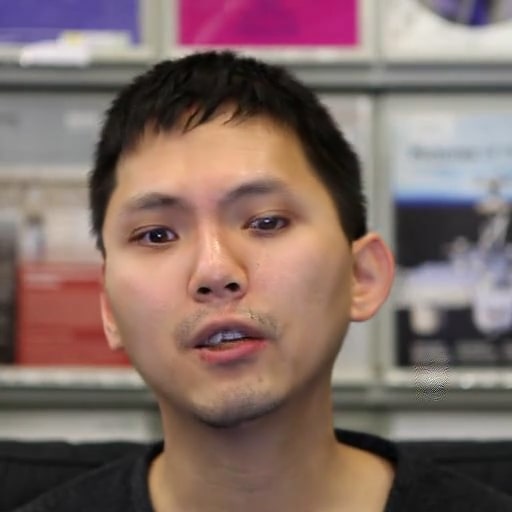}};
\node [image,right=\padding] (static-ours)     at (static-unit.east)     {\subfig{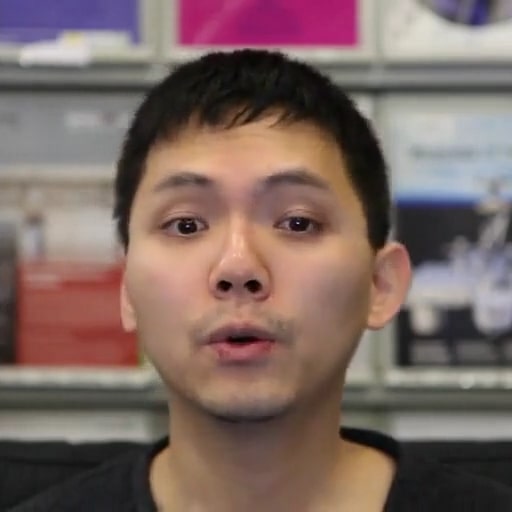}};
\filldraw [arrow,red!80!black]   (static-cyclegan.south west) ++(4.5mm,2.5mm) -- ++(3mm,3mm);
\filldraw [arrow,red!80!black]   (static-unit.south west)     ++(4.5mm,2.5mm) -- ++(3mm,3mm);
\filldraw [arrow,green!70!black] (static-ours.south west)     ++(4.5mm,2.5mm) -- ++(3mm,3mm);

\node [image,below=3pt]      (dynamic-source)   at (static-source.south)   {\subfig{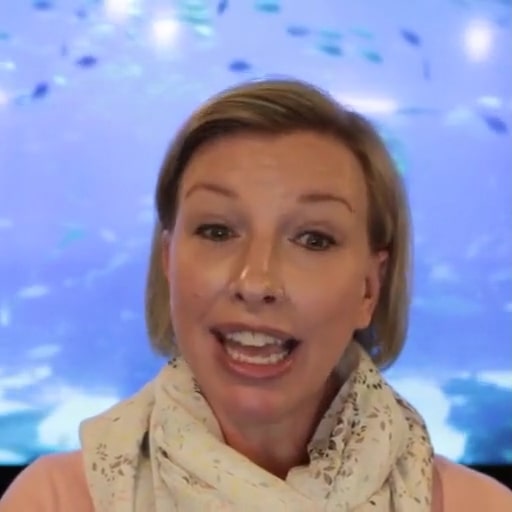}};
\node [image,right=\padding] (dynamic-cyclegan) at (dynamic-source.east)   {\subfig{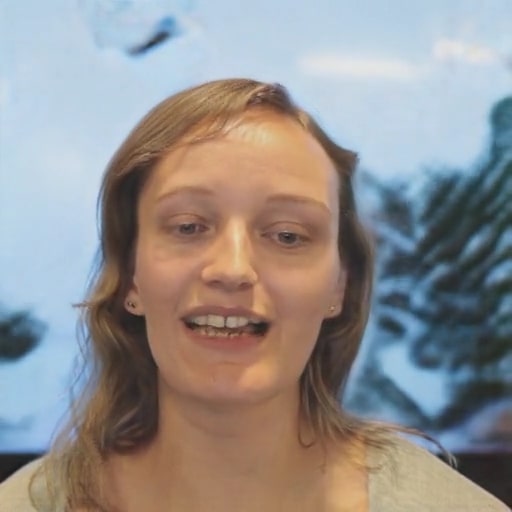}};
\node [image,right=\padding] (dynamic-unit)     at (dynamic-cyclegan.east) {\subfig{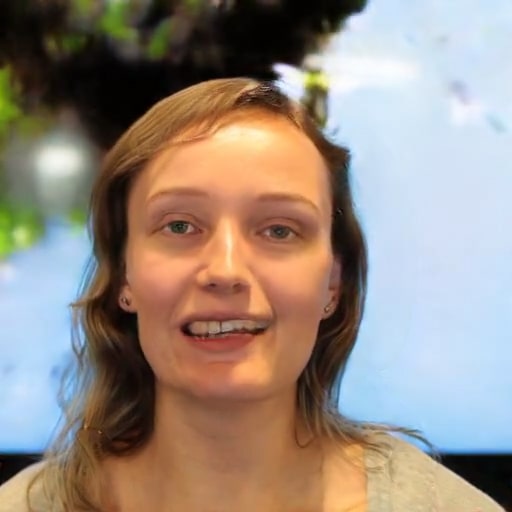}};
\node [image,right=\padding] (dynamic-ours)     at (dynamic-unit.east)     {\subfig{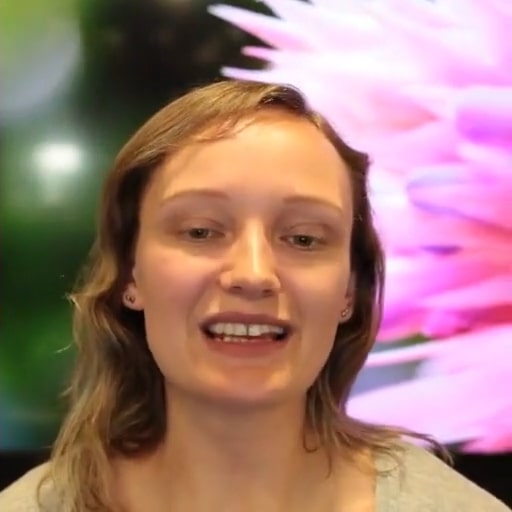}};
\filldraw [arrow,red!80!black]   (dynamic-cyclegan.south west) ++(14mm,18mm) -- ++(3mm,-3mm);
\filldraw [arrow,red!80!black]   (dynamic-unit.south west)     ++(14mm,18mm) -- ++(3mm,-3mm);
\filldraw [arrow,green!70!black] (dynamic-ours.south west)     ++(14mm,18mm) -- ++(3mm,-3mm);

\node [rowlabel] at (static-source.west)  {Static\\[-3pt]Background};
\node [rowlabel] at (dynamic-source.west) {Dynamic\\[-3pt]Background};

\node [collabel] at (static-source.north)   {Source};
\node [collabel] at (static-cyclegan.north) {CycleGAN \citeyearpar{ZhuPIE2017}};
\node [collabel] at (static-unit.north)     {UNIT \citeyearpar{LiuBK2017}};
\node [collabel] at (static-ours.north)     {Ours};

\end{tikzpicture}%
	\caption{\label{fig:comparsion_cyclegan}%
		Comparison to the unpaired image-to-image translation techniques CycleGAN \citep{ZhuPIE2017} and UNIT \citep{LiuBK2017}.
		Unsupervised transfer is a much harder problem in the image domain, due to the higher dimensionality of the problem.
		We formulate the unsupervised transfer problem in parameter space, which leads to higher quality results.
	}
\end{figure}
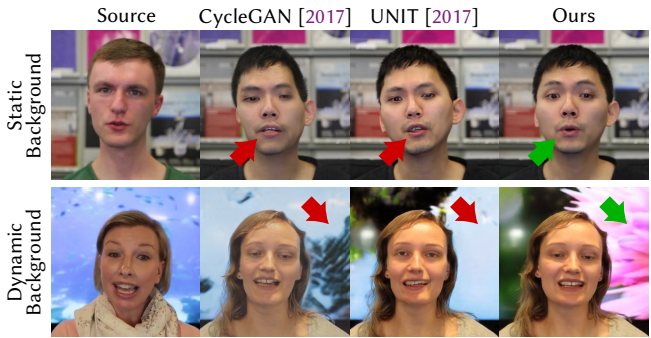

Finally, we perform a baseline comparison to CycleGAN \citep{ZhuPIE2017} and UNIT \citep{LiuBK2017}, which are unsupervised image-to-image translation techniques.
These image-based techniques are not able to fully disentangle styles from lip motion and dynamic backgrounds.
Unsupervised translation is a much harder problem in the image domain, due to the higher dimensionality of the problem.
This problem is solved by our approach that transfers the unsupervised learning problem into parameter space.
In this space, the problem is lower dimensional and the mapping can be found more robustly, as illustrated in \cref{fig:comparsion_cyclegan}.

\subsection{Ablation Study}

\begin{figure}[t]
	\centering
\begin{tikzpicture}[
image/.style={inner sep=0pt, outer sep=0pt},
collabel/.style={above=9pt, anchor=north, inner ysep=0pt, scale=0.9, align=center},
rowlabel/.style={left=17pt, rotate=90, anchor=north, inner ysep=0pt, scale=0.9, align=center},
subcaption/.style={inner xsep=0.75mm, inner ysep=0.75mm, below right},
arrow/.style={-{Latex[length=2.5mm,width=4mm]}, line width=2mm},
]

\def\padding{0pt}
\newcommand{\subfig}[2][0px 0px 0px 0px]{\includegraphics[width=0.31\linewidth, trim=#1, clip]{ablation-v2/franziska-neutral-eng-krista-smile-ger-ablation-#2.jpg}}

\node [image]                (lstm-source)    at (0,0)                {\subfig{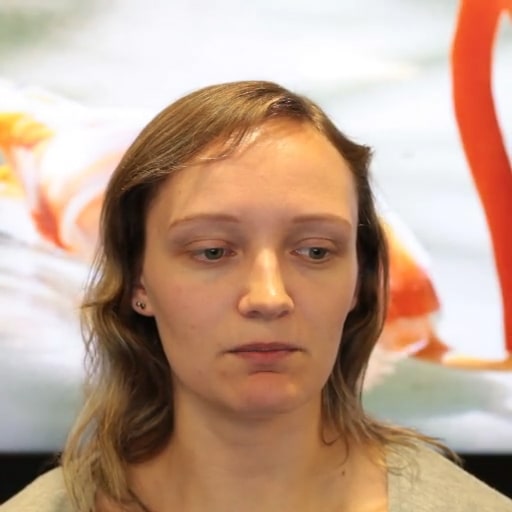}};
\node [image,right=\padding] (lstm-wo)        at (lstm-source.east)   {\subfig{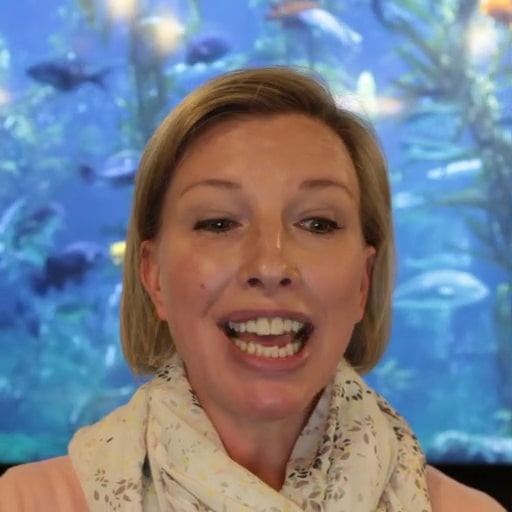}};
\node [image,right=\padding] (lstm-full)      at (lstm-wo.east)       {\subfig{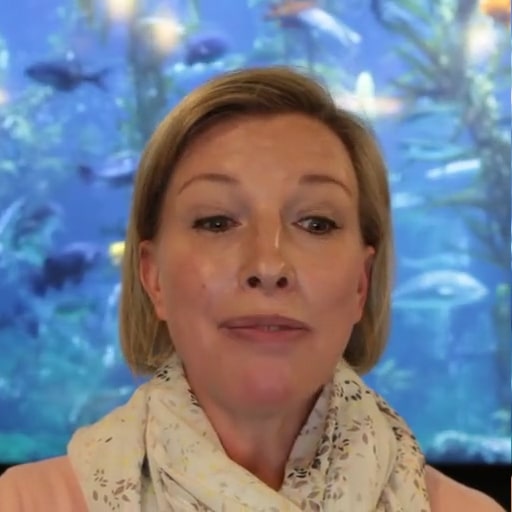}};

\node [image,below=3pt] (cycle-source)   at (lstm-source.south)  {\subfig{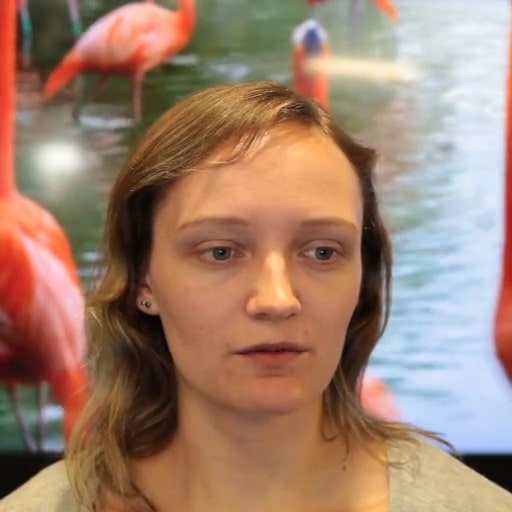}};
\node [image,below=3pt] (cycle-wo)       at (lstm-wo.south)      {\subfig{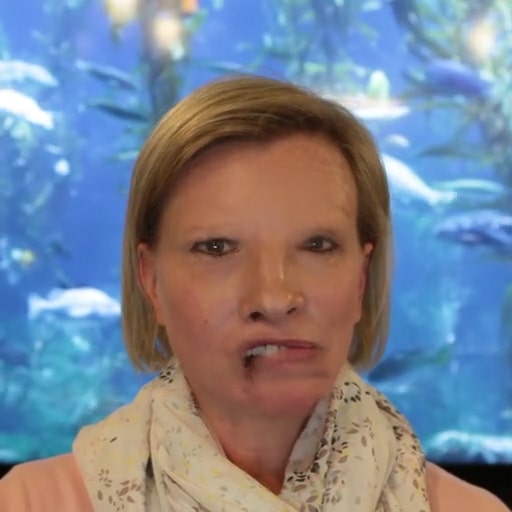}};
\node [image,below=3pt] (cycle-full)     at (lstm-full.south)    {\subfig{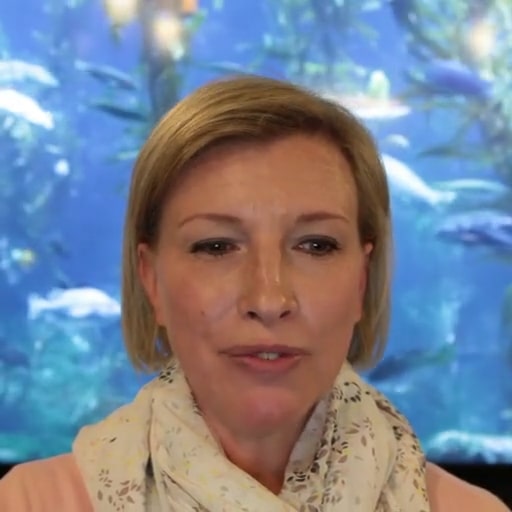}};

\node [image,below=3pt] (cosine-source)  at (cycle-source.south) {\subfig{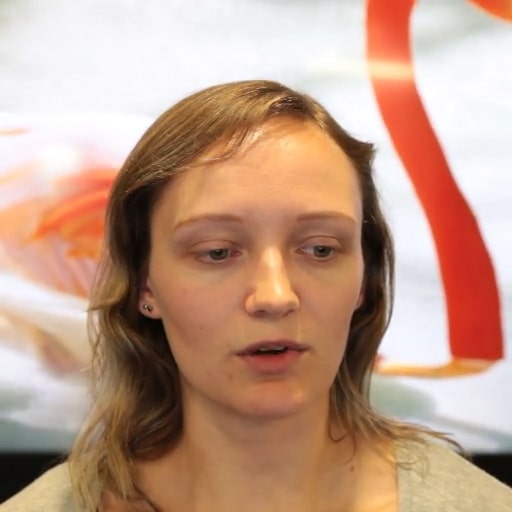}};
\node [image,below=3pt] (cosine-wo)      at (cycle-wo.south)     {\subfig{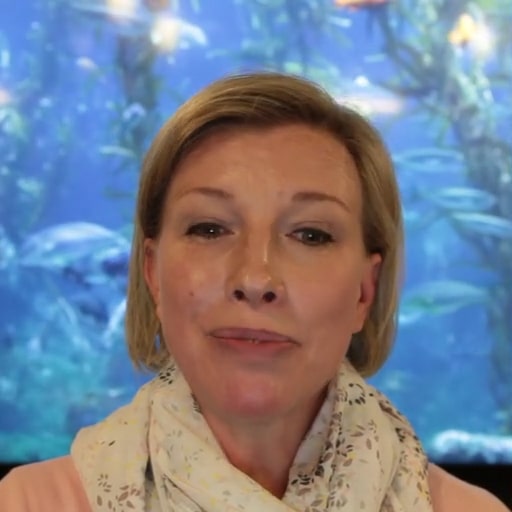}};
\node [image,below=3pt] (cosine-full)    at (cycle-full.south)   {\subfig{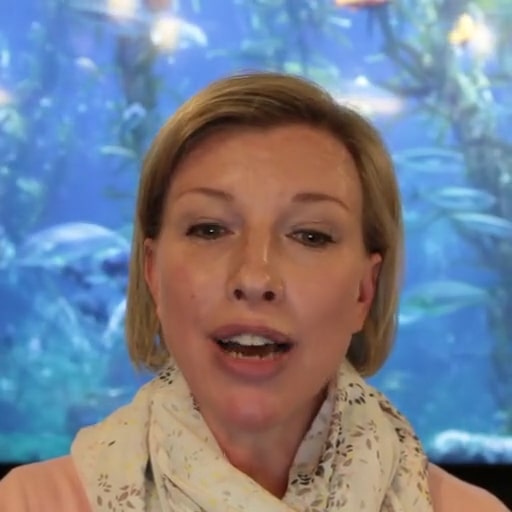}};

\node [image,below=3pt] (mask-source)  at (cosine-source.south) {\subfig{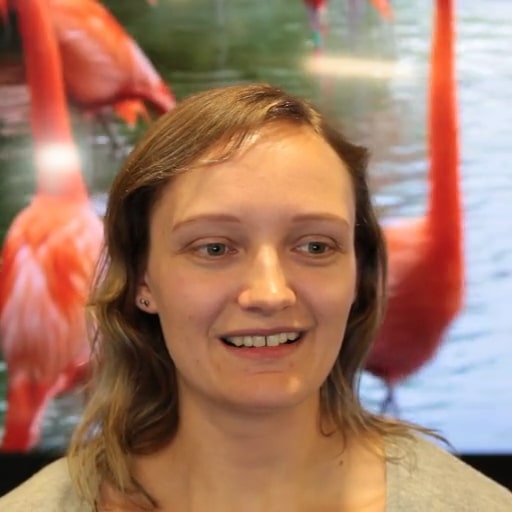}};
\node [image,below=3pt] (mask-wo)      at (cosine-wo.south)     {\subfig{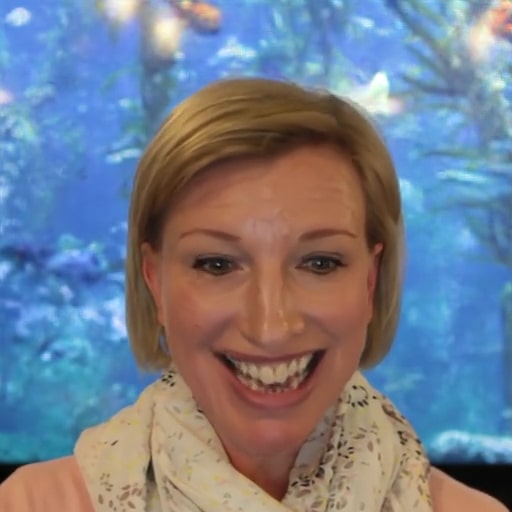}};
\node [image,below=3pt] (mask-full)    at (cosine-full.south)   {\subfig{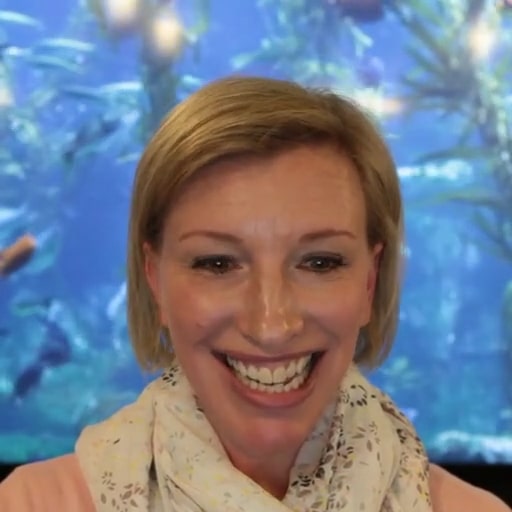}};

\node [rowlabel] at (lstm-source.west)   {Long Short-Term\\[-3pt]Memory (LSTM)};
\node [rowlabel] at (cycle-source.west)  {Cycle-Consistency\\[-3pt]Loss};
\node [rowlabel] at (cosine-source.west) {Cosine Mouth\\[-3pt]Expression Loss};
\node [rowlabel] at (mask-source.west)   {Face Attention\\[-3pt]Mask};

\node [collabel] at (lstm-source.north)   {Source};
\node [collabel] at (lstm-wo.north)       {Without};
\node [collabel] at (lstm-full.north)     {Full};

\end{tikzpicture}%
	\caption{\label{fig:ablation}%
		We perform an ablation study by disabling each component of our method to evaluate their impact on the end results.
		The LSTM unit better captures the temporal changes of facial expressions, leading to better lip-syncing and more plausible style translation.
		Without cycle consistency, the style mapping function becomes ill-posed, and thus is prone to generate invalid expressions.
		This generates significant artifacts in the final rendering.
		The cosine mouth expression loss leads to accurate lip synchronization, keeping the strength of mouth expressions close to the target actor style.
		The face mask in our neural renderer improves the visual quality of synthesized outputs, enabling dynamic backgrounds and artifact-free teeth reconstruction.
		The improvements are best visible in our video.
	}
\end{figure}
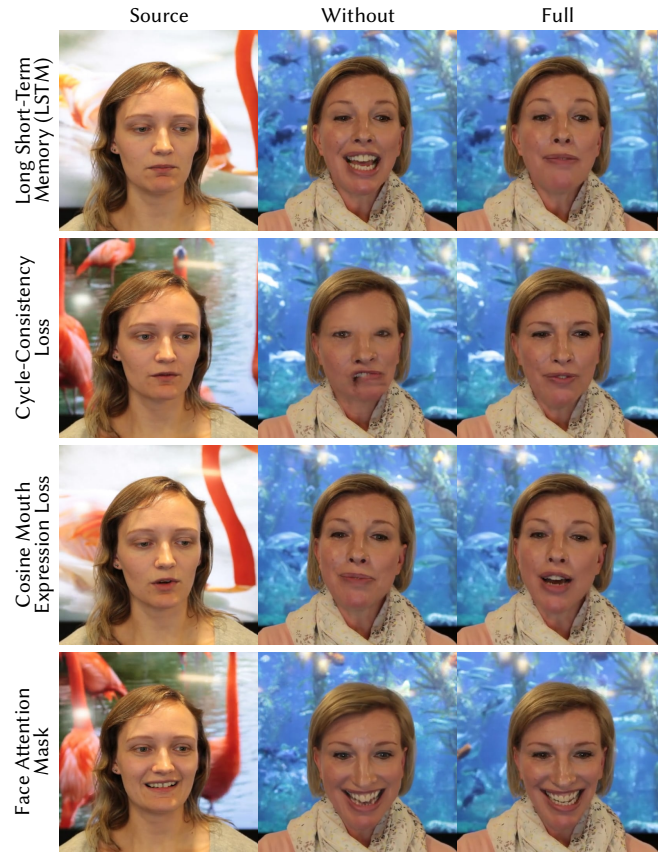

We also performed an ablation study to evaluate the components of our novel style-preserving visual dubbing approach.
A quantitative ablation study is challenging to perform, since there is no ground-truth metric available for style-preserving dubbing.
Therefore, we perform a qualitative ablation study in \cref{fig:ablation} and in the supplemental video.
We evaluate the influence of the LSTM unit, the cycle-consistency loss, the cosine mouth expression loss, and the face attention map used during training.
\cref{fig:ablation} shows that the LSTM is important in producing temporally coherent results with better lip syncing.
removing the cycle-consistency loss has the impact of distorting the face in a highly unnatural manner.
The cosine mouth expression loss better captures the audio-visual alignment of the mouth region.
Removing the background attention map leads to noticeable artifacts in the final generated background.
This is best seen in the supplementary video.

\subsection{User Study}
\label{sec:userstudy}

We performed an extensive web-based user study to quantitatively evaluate the quality of the generated visual dubbing results.
We conducted two experiments using a collection of 12 short video clips (3–8\,seconds): a subjective rating task and a pairwise comparison task using two-alternative forced choice (2AFC).
We prepared all results at 512$\times$512\,pixels resolution and showed either one or two videos side by side, depending on the experiment, but always with the target audio track.
Most video clips are from videos we recorded ourselves (e.g. see \cref{fig:result-dubbing,fig:result-exp,fig:result-similar-expressions,fig:comparsion_vdub,fig:comparsion_dvp}); they show different people talking in multiple languages with a range of facial expressions, including sarcastic, smiling and squinting.
We ensured that these clips have an unbiased distribution of source and target expressions, so as to not influence users adversely, e.g. with more favorable target expressions.
For these video clips, we compare to the ‘naïve dubbing’ baseline of combining the source audio track with the target video track without modifying the video, as well as our approach without the style translation network.
We also used a professionally dubbed video created by \citet{GarriVSSVPT2015}, i.e., a studio actor's text and speed was optimally aligned to the existing video.
We compare to both the professional dubbing and \citeauthor{GarriVSSVPT2015}'s approach.
We recruited 50 anonymous participants who took on average 13.7 minutes to complete our study.

\begin{table}[t]\small
	\caption{\label{tab:ratings}%
		User study results ($n \!=\! 50$) in response to the statement “This video clip looks natural to me”, from “––” (\textit{strongly disagree}) to “++” (\textit{strongly agree}).
		\textbf{Top:} Mean of 9 video clips with a variety of source and target actors.
		\textbf{Bottom:} Mean of 3 video clips from a professionally dubbed video.
	}
	\renewcommand*{\arraystretch}{1.2}
	\begin{tabular}{l@{\hspace{10pt}}@{\hspace{6pt}}r*{4}{@{\hspace{4pt}}r}@{\hspace{8pt}}c@{\hspace{6pt}}}
		                                & –– &  – &  o &  + & ++ & agree          \\ \hline

		Naïve dubbing                   & 54 & 33 &  5 &  8 &  0 & \phantom{0}8\% \\
		Ours without style translation  &  9 & 25 & 12 & 36 & 18 &           54\% \\
		Our with style translation      &  5 & 23 & 11 & 44 & 16 &           60\% \\ \hline

		\citet{GarriVSSVPT2015}         & 34 & 27 &  8 & 25 &  5 & 31\% \\
		Ours without style translation  & 10 & 32 & 15 & 33 &  9 & 43\% \\
		Our with style translation      &  3 & 33 & 11 & 44 &  9 & 53\% \\
		Professional dubbing            &  9 & 27 & 10 & 44 & 10 & 54\%
	\end{tabular}
\\[1.5em]
	\caption{\label{tab:comparison-naive}%
		User study results for two-alternative forced choice (2AFC) on 9 video clips ($n \!=\! 50$).
		Row “A > B” shows \%users who found A more natural.
	}
	\renewcommand*{\arraystretch}{1.2}
	\begin{tabular}{l@{\hspace{10pt}}r*{8}{@{\hspace{4pt}}r}@{\hspace{4pt}}c}
		                              &  1 &  2 &  3 &  4 &  5 &  6 &  7 &  8 &  9 & mean   \\ \hline
		
		Ours > w/o style translation  & 52 & 50 & 50 & 84 & 82 & 50 & 38 & 52 & 52 & 57\% \\
		Ours > naïve dubbing          & 96 & 84 & 92 & 94 & 92 & 98 & 98 & 74 & 76 & 89\%
	\end{tabular} 
\\[1.5em]
	\caption{\label{tab:comparison-professional}%
		User study results for two-alternative forced choice (2AFC) for a professionally dubbed video ($n \!=\! 50$, mean of 3 clips).
	}
	\renewcommand*{\arraystretch}{1.2}
	\begin{tabular}{l@{\hspace{10pt}}r@{\hspace{6pt}}r*{3}{@{\hspace{4pt}}r}@{\hspace{6pt}}}
		\textit{\% row preferred over column} & & VDub & w/o & ours & prof \\ \hline
		
		\citet{GarriVSSVPT2015}         & VDub  &  — & 39 & 27 & 21 \\
		Ours without style translation  & w/o   & 61 &  — & 22 & 26 \\
		Our with style translation      & ours  & 73 & 78 &  — & 41 \\
		Professional dubbing            & prof  & 79 & 74 & 59 &  —
	\end{tabular} 
\end{table}

The results in \cref{tab:ratings} show that our approach (60/53\% ‘natural’ rating) clearly outperforms naïve dubbing (8\%) as well as our approach without style translation (54/43\%), and it achieved the quality of professional dubbing (54\%), all while being fully automatic.
The pairwise comparison results in \cref{tab:comparison-naive} provide three insights:
1.~There is a clear preference (>80\%) of our style-preserving dubbing approach on video clips 4 and 5, in which a neutral source actor dubs a smiling target actor (shown in \cref{fig:teaser}), so the lack of style preservation is immediately obvious.
2.~Preferences for the other video clips are mixed (49\% mean), perhaps as the target actor style was not shown for comparison.
3.~While our approach is clearly preferred over naïve dubbing (89\% mean preference), there are 2 minor outliers (<80\%) where lip motions somewhat aligned to the audio by chance.
Finally, we performed pairwise comparisons on \citeauthor{GarriVSSVPT2015}'s professionally dubbed video in \cref{tab:comparison-professional}, which shows that users found our style-preserving dubbing result more natural than VDub \citeyearpar{GarriVSSVPT2015} (73\%) and our approach without style preservation (78\%).
41\% of users preferred our result to the professional dubbing result, compared to only 26\% for without style translation and 21\% for \citet{GarriVSSVPT2015}.
This is a clear improvement over the state of the art.

To better understand \emph{why} participants found our results with style translation more natural, we considered the impact of potential discrepancies between source and target actor performances, expressions and styles.
The preference for our results is not simply due to a more favorable target style:
both source and target videos of the professionally dubbed video (\cref{tab:comparison-professional}) are neutral, yet users still prefer style translation 78\% of the time.
The remaining 9 video clips are 2$\times$ smile-to-smile, 3$\times$ neutral-to-smile, 2$\times$ neutral-to-sarcastic, 2$\times$ neutral-to-squint, which is unbiased overall.
The target training video is also naturally more representative of the target actor's expressions and style, and might lack specific source expressions, such as a broad smile.
This favors our style translation approach.
In dubbing applications, the source and target videos are roughly aligned in time, which avoids visually incoherent source expression and target pose in most cases.
In practice, we observed that our approach can still cope with small misalignments and produces good results.

\subsection{Style Interpolation}

\begin{figure}
	\centering
\begin{tikzpicture}[
image/.style={inner sep=0pt, outer sep=0pt},
collabel/.style={above=9pt, anchor=north, inner ysep=0pt, scale=0.9, align=center},
rowlabel/.style={left=17pt, rotate=90, anchor=north, inner ysep=0pt, scale=0.9, align=center},
subcaption/.style={inner xsep=0.75mm, inner ysep=0.75mm, below right},
arrow/.style={-{Latex[length=2.5mm,width=4mm]}, line width=2mm},
]

\def\padding{0pt}
\newcommand{\subfig}[2][0px 0px 0px 0px]{\includegraphics[width=0.25\linewidth, trim=#1, clip]{interpolation-v2/interpolation-franziska-neutral-eng-krista-smile-ger-470-interpolation-#2.jpg}}

\node [image]                (target-00)   at (0,0)            {\subfig{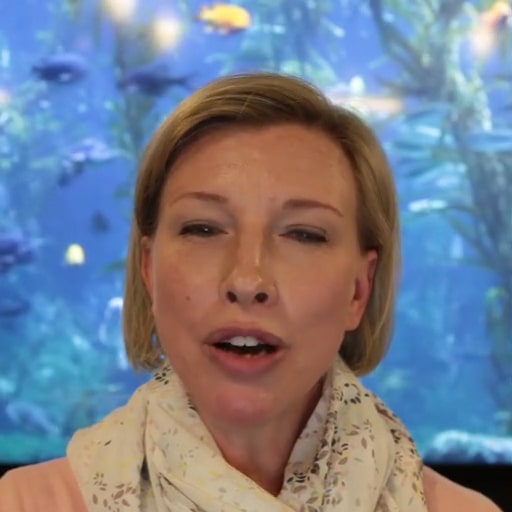}};
\node [image,right=\padding] (target-03)   at (target-00.east) {\subfig{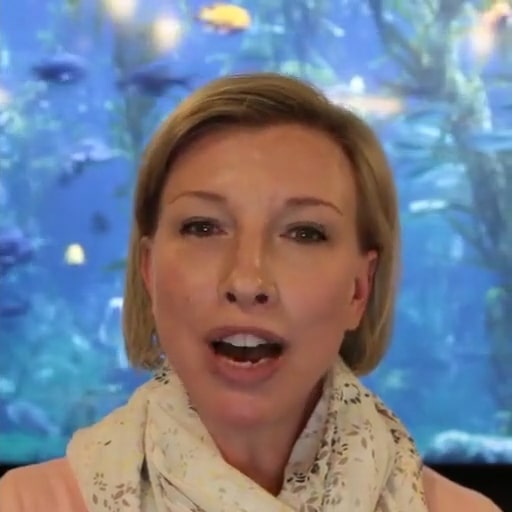}};
\node [image,right=\padding] (target-07)   at (target-03.east) {\subfig{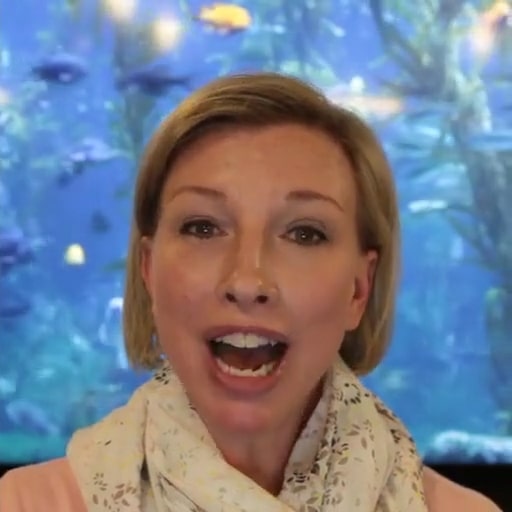}};
\node [image,right=\padding] (target-10)   at (target-07.east) {\subfig{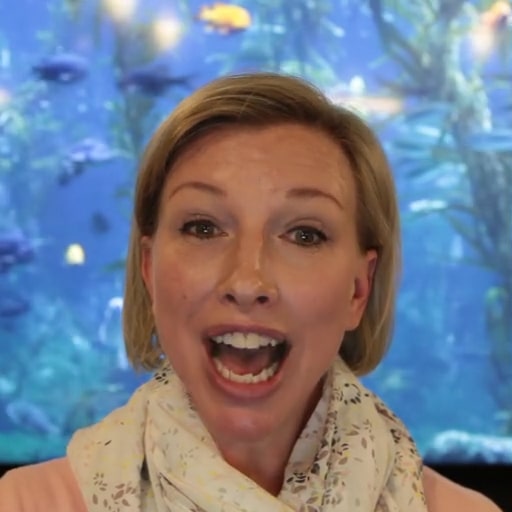}};

\node [collabel,below=3pt] at (target-00.south) {$\alpha=$ 0.0\\[-3pt](Source Style)};
\node [collabel,below=3pt] at (target-03.south) {0.3};
\node [collabel,below=3pt] at (target-07.south) {0.7};
\node [collabel,below=3pt] at (target-10.south) {1.0\\[-3pt](Target Style)};

\end{tikzpicture}%
	\caption{\label{fig:result_interpolation}%
		Seamless interpolation between the source and target actor style.
        We linearly blend the original and translated source expressions using the weight $\alpha \!\in\! [0, 1]$, and visualize the result with our neural renderer.
        Note that the style interpolation is achieved with synchronized lip motions.
	}
\end{figure}
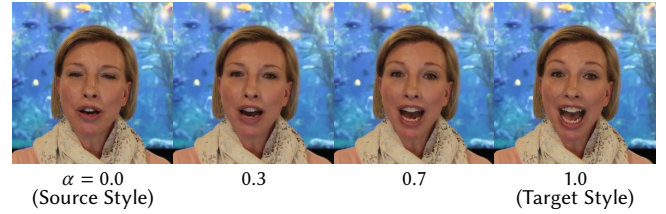

In addition to style-preserving visual dubbing, our approach can also be used to change the style of the target actor in interesting and meaningful ways.
For example, we can smoothly blend between the style of the source and the target actor, see \cref{fig:result_interpolation}, while controlling the target sequence.
This can, for example, be used to make the target a bit more happy, neutral or sad, by mixing in some of the style of the actor in the source sequence.
One could imagine using this technique as a postprocessing step in a movie production to slightly adjust the tone of an already recorded performance.


\section{Discussion}

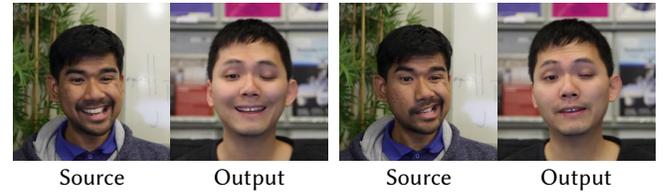
\begin{figure}
	\centering
\begin{tikzpicture}[
image/.style={inner sep=0pt, outer sep=0pt},
collabel/.style={below=-3mm, align=center},
rowlabel/.style={left=10pt, rotate=90, anchor=north, inner ysep=0pt},
subcaption/.style={inner xsep=0.75mm, inner ysep=0.75mm, below right},
]

\def\padding{0pt}
\newcommand{\subfig}[2][0px 0px 0px 0px]{\includegraphics[width=0.245\linewidth, trim=#1, clip]{limitation-v2/ikhsanul-smile-jiayi-sarcastic-#2.jpg}}

\node [image]                (source1) at (0,0)          {\subfig{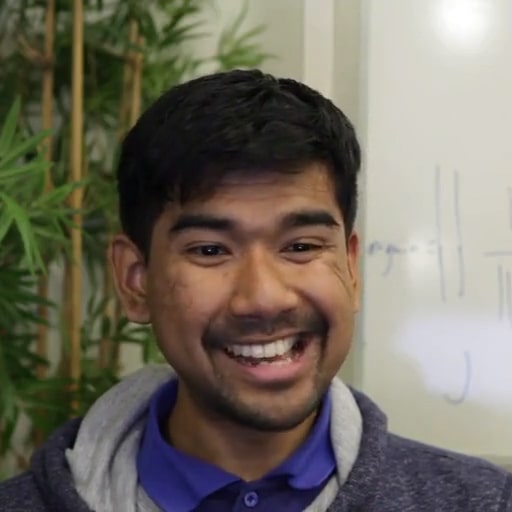}};
\node [image,right=\padding] (ours1)   at (source1.east) {\subfig{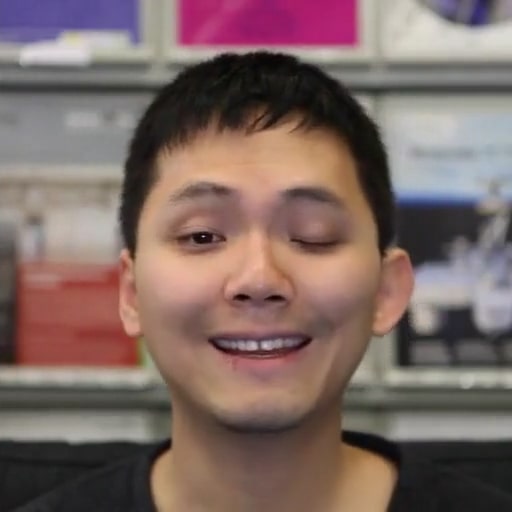}};

\node [image,right=\padding] (source2) at ([xshift=0.5em] ours1.east) {\subfig{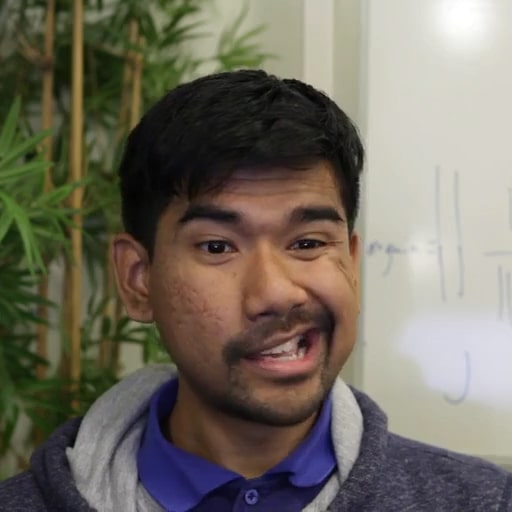}};
\node [image,right=\padding] (ours2)   at (source2.east)              {\subfig{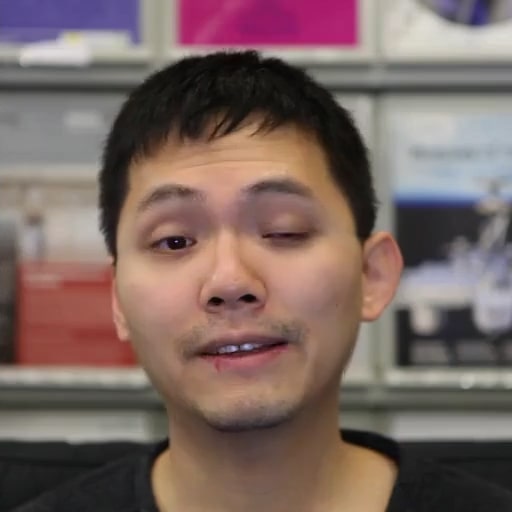}};

\node [collabel,below=0pt] at (source1.south) {Source};
\node [collabel,below=0pt] at (ours1.south)   {Output};
\node [collabel,below=0pt] at (source2.south) {Source};
\node [collabel,below=0pt] at (ours2.south)   {Output};
\end{tikzpicture}%
	\caption{%
		Our method might generate implausible facial expressions such as one-eyed twitching when the distributions of the facial expressions of the source and target domains are far apart.
		This could be avoided by incorporating additional constraints concerning facial anatomy.
		The temporal artifact is visible in the supplemental video.
	}
	\label{fig:limitation}
\end{figure}

In this work, we have demonstrated high-quality style-preserving visual dubbing results for a large variety of sequences. 
Our approach makes a step towards further simplifying localization of media, in particular movies and TV content, to different countries and languages.  
Nevertheless, our approach has a few limitations that can be tackled in the future.

Monocular face reconstruction is an extremely challenging problem and may fail for extreme illumination conditions or head poses, such as are often observed in two person dialogue shots.
In these cases, the facial expressions can not be robustly recovered and thus the neural face renderer can not be reliably trained.
More robust face reconstruction techniques could alleviate this problem in the future.
Similar to many other data-driven techniques, training our neural face renderer requires a sufficiently large training corpus.
Generalizing across subjects, to enable dubbing in settings where only a short video clip of the target actor is available, is an open challenge.
Related to this, our approach works well inside the span of the training corpus, but generalization to unseen expressions is hard, e.g., synthesizing an extreme opening of the mouth, if this has not been observed before, might lead to visual artifacts.

The unsupervised training of the style translation might generate implausible facial expressions if the source and target distributions are too different.
We noticed this is mostly the case when the source dubber has an extreme style, such as squinting or twitching.
In \cref{fig:limitation}, the extreme twitching of the source actor produced visual artifacts in the dubbed result.
These artifacts could potentially be reduced by incorporating additional constraints on facial anatomy during training.
Out of 19 videos in our dataset (\cref{tab:datasets}), only 2 contain such extreme styles (11\%).
Differences in ethnicity or gender between the source and target may further contribute to differences in styles, which we leave for future work.
We acknowledge minor visual artifacts in the final renderings but consider improvements in the rendering quality orthogonal to our main contribution of style-preserving dubbing.


\section{Conclusion}
We have presented the first style-preserving visual dubbing approach that maintains the signature style of the target actor including their person-specific idiosyncrasies.
At the core of our approach is an unpaired temporal parameter-to-parameter translation network that can be trained in an unsupervised manner using cycle-consistency and mouth expression losses.
Afterwards, photorealistic video frames are synthesized using a layered neural face renderer.
Our results demonstrate that a large variety of source and target expressions, across subjects from different ethnicities and speaking different languages, can be handled well.
We see our approach as a step towards solving the important problem of video dubbing and we hope it will inspire more research in this direction.

\begin{acks}
We are grateful to all our actors and the reviewers for their valuable feedback.
We thank True-VisionSolutions Pty Ltd for kindly providing the 2D face tracker and Adobe for a Premiere Pro CC license.
This work was supported by
\grantsponsor{4DRepLy}{ERC}{https://erc.europa.eu/funding/consolidator-grants} \grantnum{4DRepLy}{Consolidator Grant 4DRepLy (770784)},
the \grantsponsor{MPCVCC}{Max Planck Center for Visual Computing and Communications (MPC-VCC)}{https://www.mpc-vcc.org/},
\grantsponsor{CAMERA}{RCUK}{https://www.ukri.org/} grant \grantnum{CAMERA}{CAMERA (EP/M023281/1)},
and an \grantsponsor{Fellowship}{EPSRC-UKRI}{https://epsrc.ukri.org/} \grantnum{Fellowship}{Innovation Fellowship (EP/S001050/1)}.
\end{acks}

\bibliographystyle{ACM-Reference-Format}
\bibliography{FaceStyleTransfer}


\appendix
\section{Appendix}

This appendix lists the used datasets in \cref{tab:datasets}.

\begin{table}[b]
	\newcommand{\subfig}[2][0px 0px 0px 0px]{\raisebox{-.3\height}{\includegraphics[width=6mm, trim=#1, clip]{datasets/#2.jpg}}}%
	\caption{\label{tab:datasets}%
		List of datasets used in our results and comparisons as source and/or target videos.
		The language is given using ISO 639-2/B codes.
		Obama image courtesy of the White House (public domain).
	}
	\renewcommand*{\arraystretch}{1.2}
	\begin{tabular}{clclr}
		            Image              & Name   & Language & Style           & \#Frames \\ \hline
		  \subfig{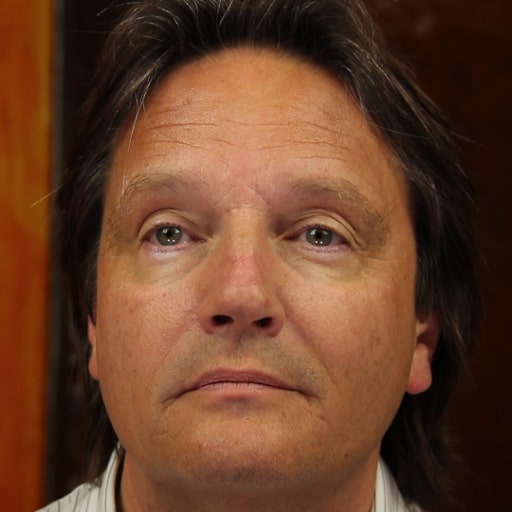}   & David  &   eng    & neutral + smile &    3,232 \\
		   \subfig{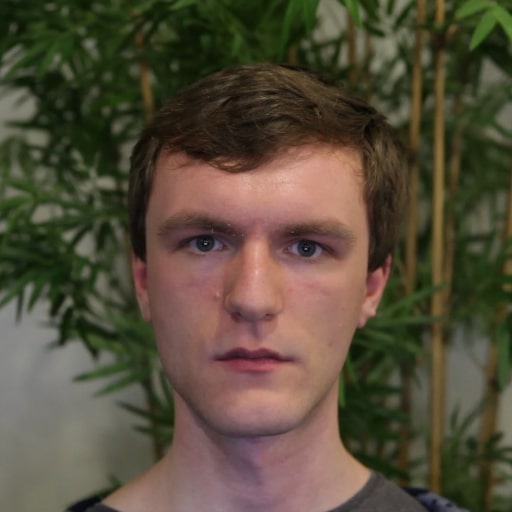}    & E.     &   ger    & angry           &    8,628 \\
		  \subfig{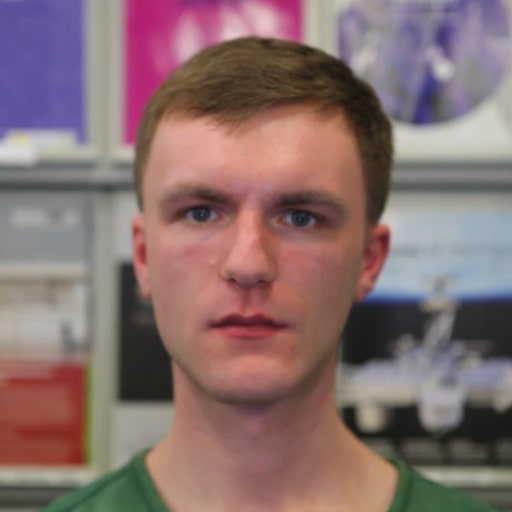}   & E.     &   ger    & neutral         &    8,222 \\
		  \subfig{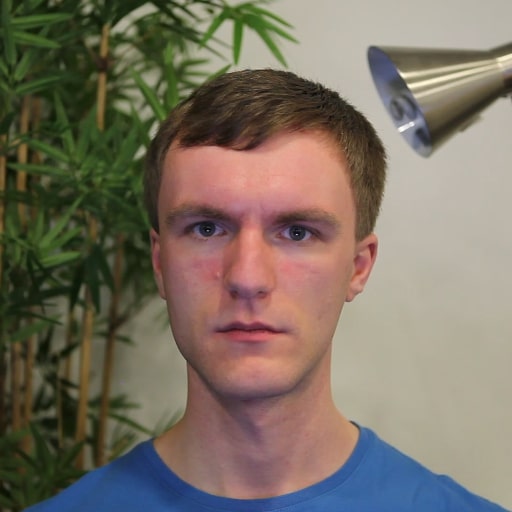}    & E.     &   ger    & squint          &    7,711 \\
		\subfig{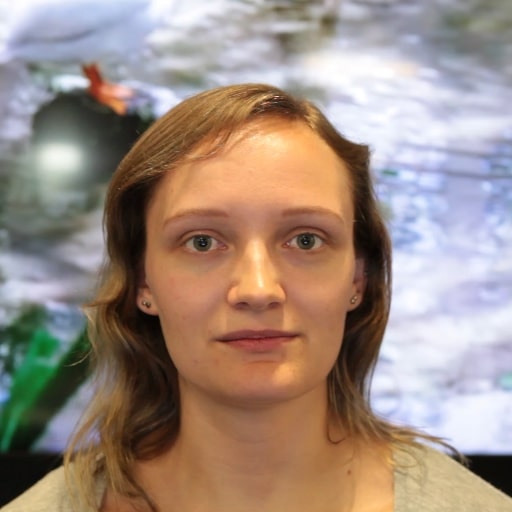} & F.     &   eng    & neutral         &    8,020 \\
		 \subfig{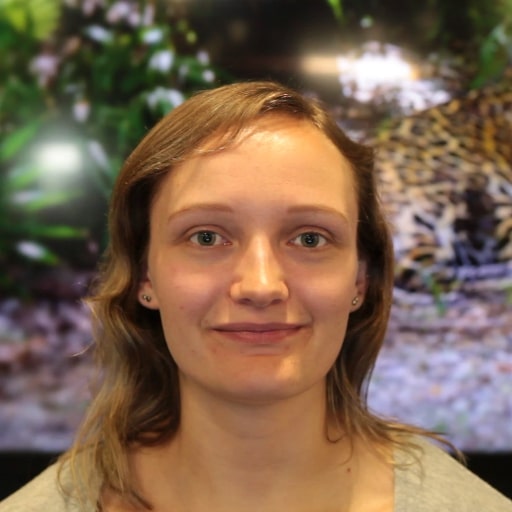}  & F.     &   ger    & smile           &    8,333 \\
		 \subfig{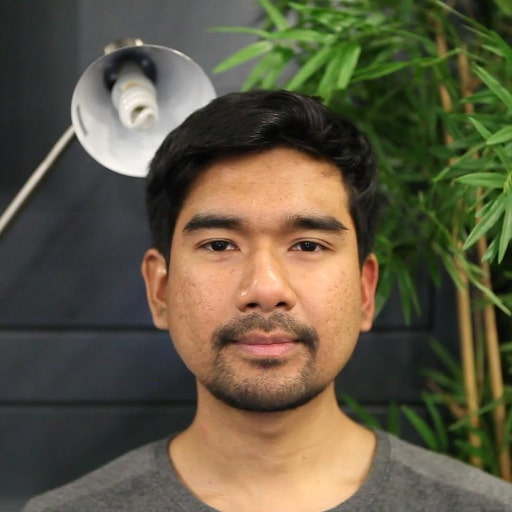}   & I.     &   ind    & smile           &    8,782 \\
		 \subfig{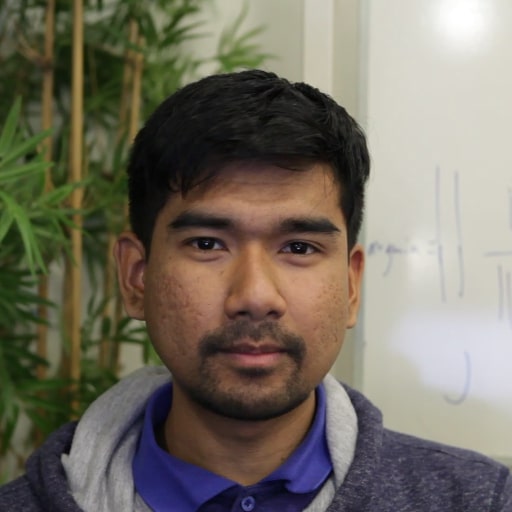}   & I.     &   eng    & smile           &   10,138 \\
		  \subfig{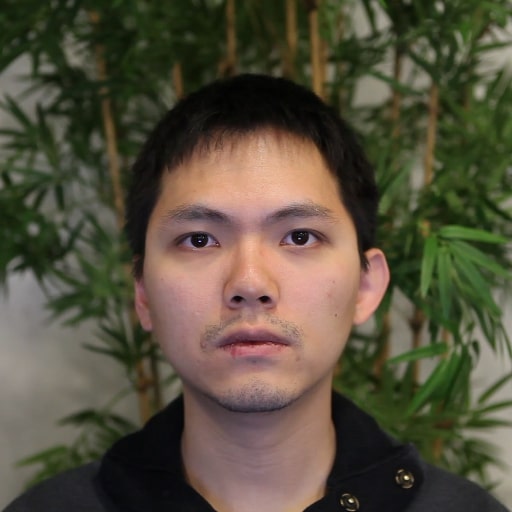}   & J.     &   eng    & neutral         &    9,070 \\
		 \subfig{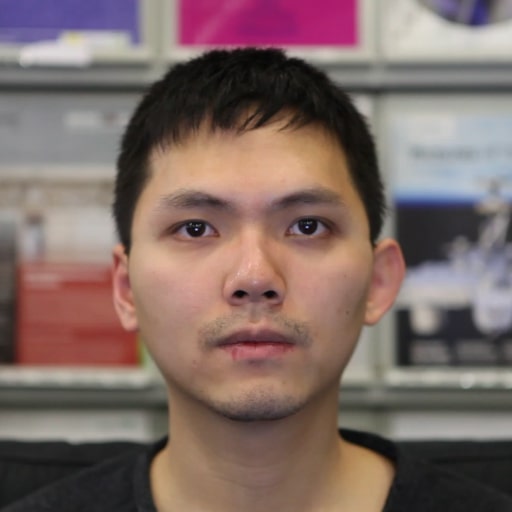}  & J.     &   eng    & sarcastic       &    8,977 \\
		  \subfig{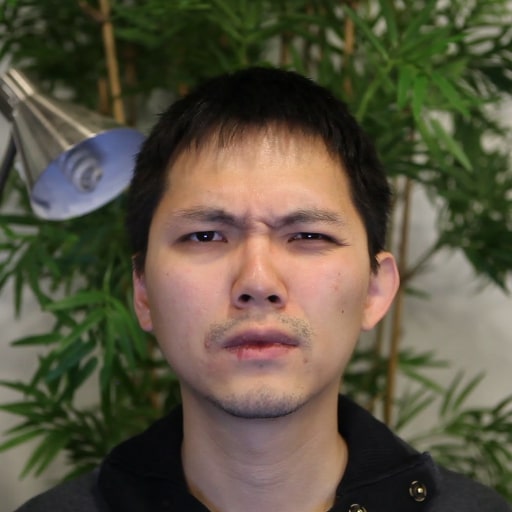}    & J.     &   eng    & squint          &    8,048 \\
		  \subfig{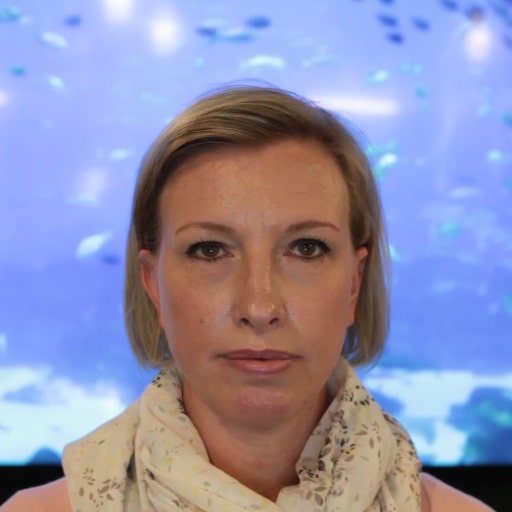}    & K.     &   eng    & smile           &    9,972 \\
		  \subfig{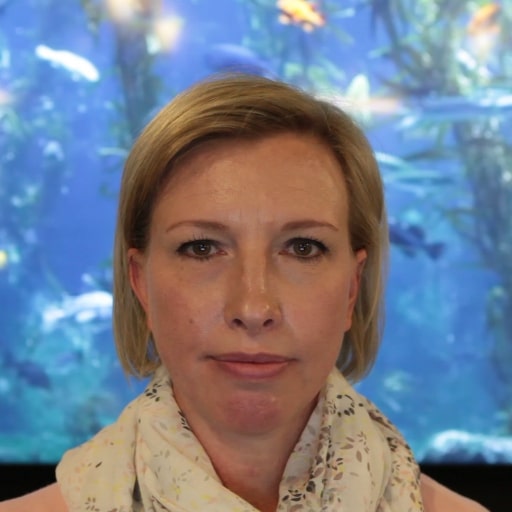}    & K.     &   ger    & smile           &   11,745 \\
		 \subfig{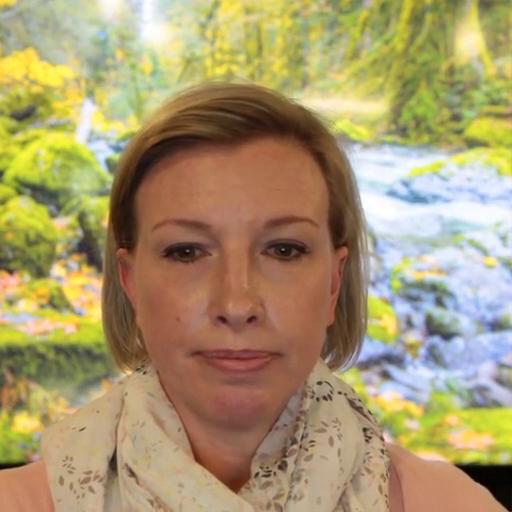}   & K.     &   eng    & neutral         &    9,664 \\
		\subfig{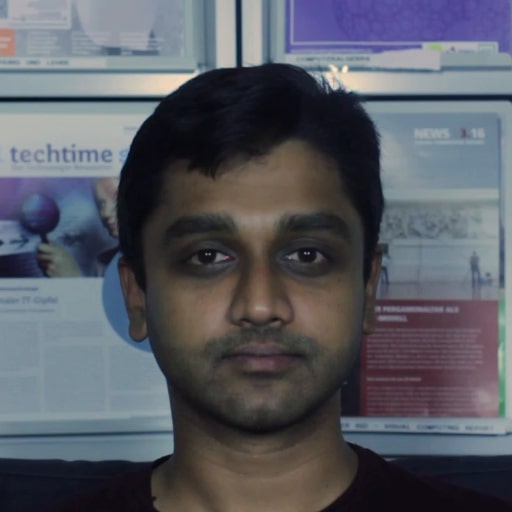} & M.     &   eng    & smile           &    8,911 \\
		\subfig{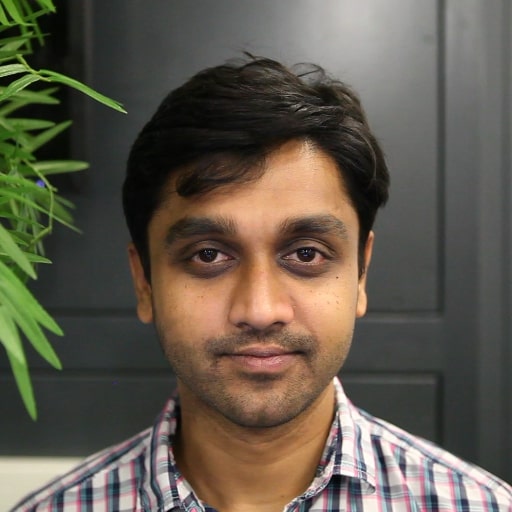} & M.     &   kan    & smile           &    8,294 \\
		 \subfig{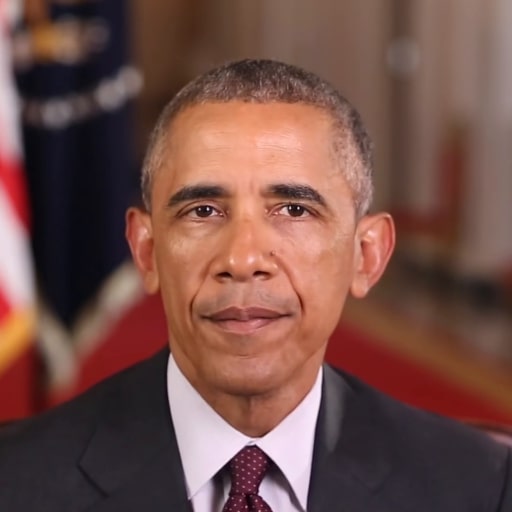}   & Obama  &   eng    & neutral         &    1,945 \\
		 \subfig{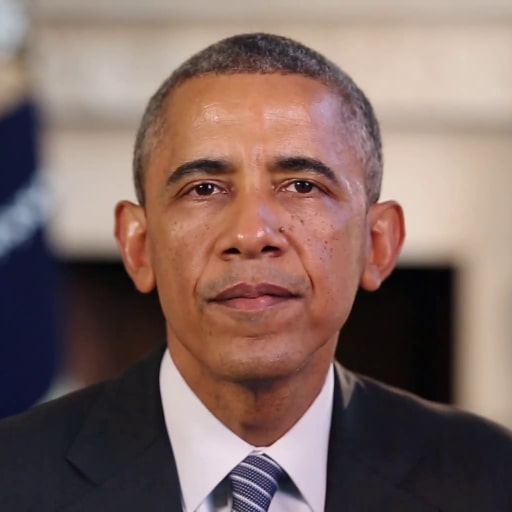}   & Obama  &   eng    & neutral         &    3,613 \\
		 \subfig{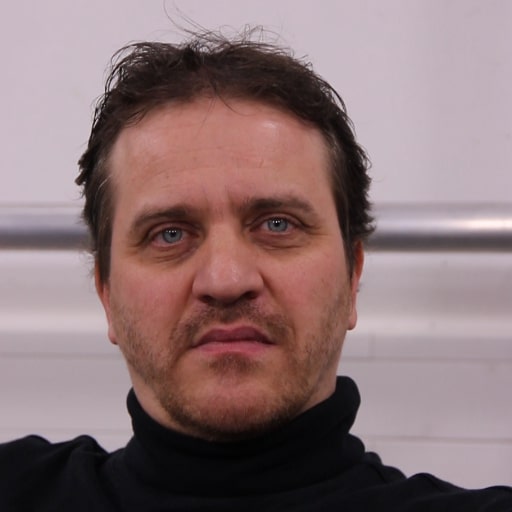}   & Thomas &   ger    & neutral + smile &    3,232 \\ \hline
	\end{tabular}
\end{table}

\end{document}